\documentclass[fleqn,10pt,final]{wlscirep}
\usepackage[utf8]{inputenc}
\usepackage[T1]{fontenc}
\usepackage{multirow}
\usepackage{placeins}
\usepackage{comment}
\usepackage{textcomp}

\title{Knowledge Graph modulated Deep Learning for Limited-Sample Clinical Data Analysis}

\author[1,**]{Yuwei Xue}
\author[1,**]{Sakib Mostafa}
\author[2,3,4]{James Zou}
\author[5]{Joseph Liao}
\author[1,6,7]{Maximilian Diehn}
\author[8]{Ash A. Alizadeh}
\author[1,2,9,*]{Lei Xing}
\author[1,*]{Md Tauhidul Islam}

\affil[1]{Department of Radiation Oncology, Stanford University, Stanford, California, USA}
\affil[2]{Department of Electrical Engineering, Stanford University, Stanford, California, USA}
\affil[3]{Department of Biomedical Data Science, Stanford University School of Medicine, Stanford, California, USA}
\affil[4]{Department of Computer Science, Stanford University, Stanford, California, USA}
\affil[5]{Department of Urology, Stanford University School of Medicine, Stanford, California, USA}
\affil[6]{Stanford Cancer Institute, Stanford University, Stanford, California, USA}
\affil[7]{Institute for Stem Cell Biology and Regenerative Medicine, Stanford University, Stanford, California, USA}
\affil[8]{Department of Medicine, Division of Oncology, Stanford University, Stanford, California, USA}
\affil[9]{Institute of Computational and Mathematical Engineering, Stanford University, Stanford, California, USA}
\affil[**]{Co-first author}

\begin{abstract}
Biological systems are governed by structured molecular interactions, where pathways, regulatory circuits, and functional gene relationships shape cellular behavior and disease progression. Much of this critical biological knowledge is naturally represented as graphs. However, most biomedical AI models cannot directly integrate knowledge in graph form and instead require low-dimensional representations of biological knowledge. This compression often leads to information loss and substantially reduces model performance, especially in limited-sample settings frequently encountered in clinical studies. Here, we introduce Graph-in-Graph (GiG), a knowledge graph–modulated deep learning framework for data-efficient clinical prediction from limited patient samples. GiG represents each patient as a standalone modular graph, in which curated biological knowledge graphs can be integrated as edges and patient-specific information such as gene expression profiles can be integrated as node features. This flexible design enables the integration of multiple biological knowledge graphs while allowing the model to learn patient-level disease representations that preserve biologically meaningful information, including gene–gene interactions and pathway topology.
Across several patient cohorts comprising nearly 9,700 patients and five clinical tasks, including cancer detection from liquid biopsy dataset from Stanford Hospital, prostate cancer diagnosis, and a 32-class pan-cancer classification, GiG consistently outperforms traditional methods by a large margin, particularly in limited-sample settings. On the challenging prostate cancer diagnosis task, GiG improves macro-F1 performance by up to 49 percentage points when compared to state-of-the-art (SOTA) methods. Control experiments replacing real pathway graphs with random topologies confirm that the performance gains arise from biologically grounded knowledge graph structure rather than graph modeling alone. These findings show that knowledge graph–modulated deep learning can improve robustness, interpretability, and sample efficiency in clinical data analysis, providing a principled framework for integrating biological knowledge graphs into predictive modeling.
\end{abstract}

\begin{document}

\flushbottom
\maketitle

\thispagestyle{empty}

\section{Introduction}
Biological systems are organized across multiple functional levels, with molecular interactions forming gene pathways that collectively determine cellular behavior and disease phenotypes. In these systems, genes do not operate in isolation but participate in interconnected pathways, signaling cascades, structured regulatory circuits, and metabolic routes that have been mapped and recorded through decades of research in molecular and systems biology \cite{hanahan2022hallmarks, drier2013pathway}. Modern genomic assays such as bulk transcriptomic, single-cell sequencing, and liquid biopsy now measure this molecular activity at high resolution across large patient cohorts, creating opportunities to connect gene-level measurements to clinical outcomes at a scale that was previously inaccessible \cite{steyaert2023multimodal}. Whether those opportunities translate into clinically useful predictive models depends in large part on computational frameworks that incorporate the organization of biology, rather than treating transcriptomic data as an unstructured list of values to be passed through a generic model \cite{chandak2023building}.

\begin{figure}
\centering
\includegraphics[width=\linewidth]{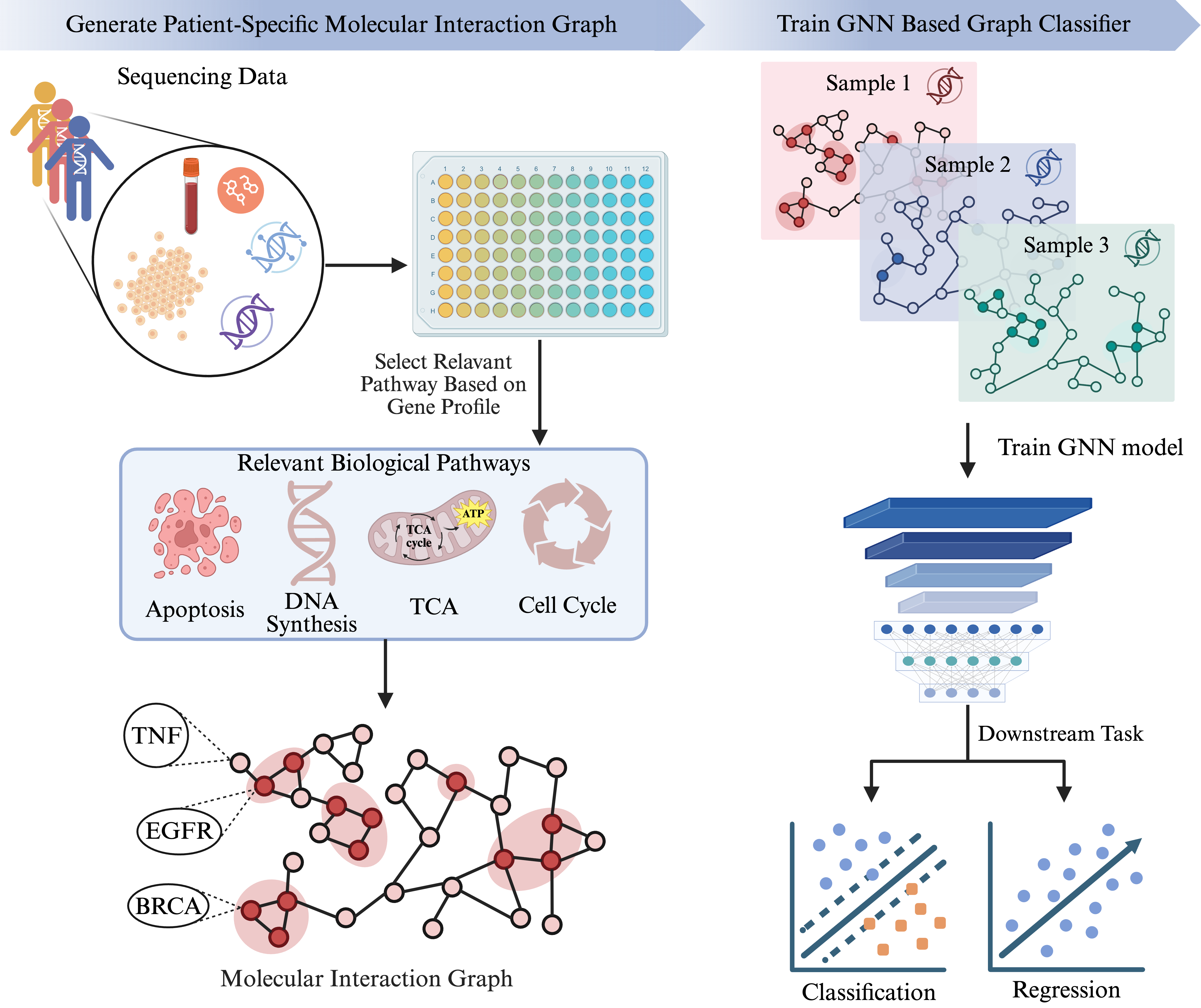}
\caption{\textbf{Workflow of the proposed GiG framework.} GiG constructs patient-specific molecular interaction graphs by integrating transcriptomic profiles with curated biological pathway knowledge. Starting from sequencing-derived gene expression data, pathways relevant to each patient are selected based on the sample-specific gene expression profile. Curated pathway interactions are then merged to generate a molecular interaction graph in which nodes represent genes and edges encode known biological relationships. These patient-specific graphs are subsequently used as inputs to a graph neural network (GNN) classifier, enabling graph-level learning across a cohort of samples. The learned graph representations can then be applied to downstream predictive tasks such as disease classification and regression.}
\label{fig:figure1}
\end{figure}

Most genomic machine learning pipelines today continue to treat expression data as a flat table \cite{hao2024large, smith2020standard}. Gene values become columns in a matrix, and classifiers search for combinations of these columns that separate disease states from healthy ones. This framing provides convenient inputs for standard deep learning architectures, yet it ignores two foundational properties of the underlying biology. First, genes participate in curated interaction networks whose topology has been established through extensive experimental validation and carries real biological meaning \cite{milacic2024reactome, szklarczyk2023string, kanehisa2023kegg}. Second, the activity of a given pathway varies substantially between patients \cite{drier2013pathway}, so a single fixed network representation cannot capture how molecular signals are distributed across individuals within a disease cohort. Graph neural network methods developed for biomedical applications partially address the first property by introducing a prior interaction graph \cite{wang2021mogonet}, but most current approaches rely on a single global network shared across all samples, or construct patient-similarity networks in which each patient is again reduced to a feature vector \cite{valous2024graph}. Neither design allows pathway topology and patient-specific expression to interact at the level of message passing.

Several bodies of work have approached parts of this problem from different angles. Pathway-informed neural architectures constrain layer connectivity to follow curated biological groupings, producing models whose internal structure reflects canonical pathway annotations \cite{elmarakeby2021biologically, hartman2023interpreting, ma2018using, lin2021using}. Graph neural networks applied to protein-protein interaction networks have shown that message passing over biological topology can improve performance on gene-level and sample-level prediction tasks \cite{jha2022prediction, maslov2002specificity, gao2023hierarchical}. Patient-similarity networks combine clinical and molecular features to propagate label information across cohorts, leveraging the observation that phenotypically similar patients often share outcomes \cite{kesimoglu2023supreme, wang2014similarity, pai2019netdx}. More recent work has extended graph foundation modeling to biomedical networks, learning transferable structural representations that can be adapted to multiple graph analysis tasks with limited labeled data \cite{huang2024foundation, hao2024large, moor2023foundation}. However, currently there are no methods, in which every patient can be represented as a pathway-structured graph of their own, with the interaction topology integrated into each sample rather than supplied as an external prior shared across the cohort.

Here to address the limitation of existing methods and integrate biological knowledges as stand-alone graphs into deep learning pipelines, we introduce Graph-in-Graph (GiG), a framework in which each patient is represented as a pathway-structured graph and downstream tasks are performed across a population of such graphs. Nodes correspond to genes drawn from the patient transcriptomic profile, and edges encode the topology of curated biological pathways sourced from WikiPathways \cite{agrawal2024wikipathways}, so that the functional organization of molecular interactions is encoded directly into the graph of every sample. Node features combine the patient gene expression values, a learnable gene-identity embedding that anchors each node to its molecular role across the cohort, and a co-expression signal derived from sample-wise correlation structure. Edge weights are set by the same correlation structure, allowing pathway topology to shape how information propagates during message passing while patient-specific expression controls what information actually travels along the edges. In this formulation, the lower-level pathway structure sits inside the higher-level patient graph, not as a separate preprocessing step but as an intrinsic component of the graph itself. Because the design is backbone agnostic, GiG can be instantiated with GCN, GraphSAGE, GIN, GAT, or any compatible graph neural network, which allows us to separate the contribution of the pathway-structured representation from the choice of message-passing architecture.

We evaluated GiG on several transcriptomic cohorts totaling close to 9,700 patients across five classification tasks, including a liquid biopsy dataset (RareSeq)\cite{nesselbush2025ultrasensitive}, a prostate cancer cohort \cite{elmarakeby2021biologically}, and a 32-class multi-omic pan-cancer benchmark drawn from TCGA \cite{liu2018integrated}. The most striking gain came from the pan-cancer classification task, which is the most challenging of the three cohorts because of the limited patient samples per class. Across 32 cancer types, GiG achieved 92\% accuracy and 88\% macro F1, while the strongest conventional graph neural network baseline operating on the same inputs reached only 63\% accuracy and 39\% F1, corresponding to a gain up to 49\% in macro F1 without any change to the underlying expression data or training protocol. The same pattern of consistent improvements over standard backbones held across the liquid biopsy and prostate cancer cohorts. To confirm that these gains come from the pathway topology itself rather than from node features alone, we replaced the real pathway graphs with Erdős Rényi and degree-preserving random controls and measured the resulting drop in classification accuracy. We also characterized the structural distinctiveness of the real graphs using graphlet orbit statistics, which showed that pathway-derived topologies occupy a region of structural feature space that is clearly separated from both random-graph families across every dataset and task.

In summary, here, we propose a per-patient graph formulation that integrates pathway topology directly into every sample, moving beyond the shared global network assumption that dominates current graph based genomic modeling. With rigorous validation, we demonstrate consistent performance gains across three heterogeneous cohorts and five classification tasks that differ in sample size, disease granularity, class cardinality, and class balance, showing that the benefits of pathway-structured modeling are not confined to a single benchmark or task type. We also validate through controlled structural perturbations and graphlet orbit analysis that the observed gains arise from the pathway topology itself, not from the expression features in isolation. Together these results position GiG as a practical foundation for predictive modeling of transcriptomic data in precision medicine, in which the organization of biology is treated not as an afterthought to a flat feature vector but as a structural component of the model.

\section{Results}
\subsection{GiG outperforms SOTA GNNs in cancer subtype and stage classification on an ultrasensitive cfRNA dataset}
We first evaluated GiG's ability to accurately detect cancer and predict cancer subtype and stage, of patient samples in the RARE-Seq cfRNA dataset  from Stanford hospital \cite{nesselbush2025ultrasensitive}. RARE-Seq is designed to detect extremely low-abundance tumor-derived transcripts in plasma. Unlike tissue-based transcriptomic datasets, where tumor signal is relatively strong and localized, cfRNA measurements reflect a highly diluted and fragmented representation of tumor biology embedded within a background of circulating RNA from multiple tissue sources \cite{vorperian2025cell, morlion2026patient, ma2024liquid}. In other words, the RARE-Seq data set is characterized by low signal-to-noise ratios, substantial inter-patient heterogeneity, and significant class imbalance. Therefore, accurate prediction of cancer presence, subtype, and stage within this dataset requires a model to extract weak, distributed signals that are often obscured by noise and variability. In this study, following traditional gene expression data pre-processing methods, we selected 10,000 highly variable genes (HVGs) out of over 60,000 genes from the RARE-Seq dataset to train and test our model. By focusing on genes that exhibit the greatest variability across samples, this setup allows us to capture biologically meaningful differences between disease states while reducing the influence of low-variance, noise-dominated features.

\begin{figure}[p]
\centering
\includegraphics[width=\linewidth,height=\textheight,keepaspectratio]{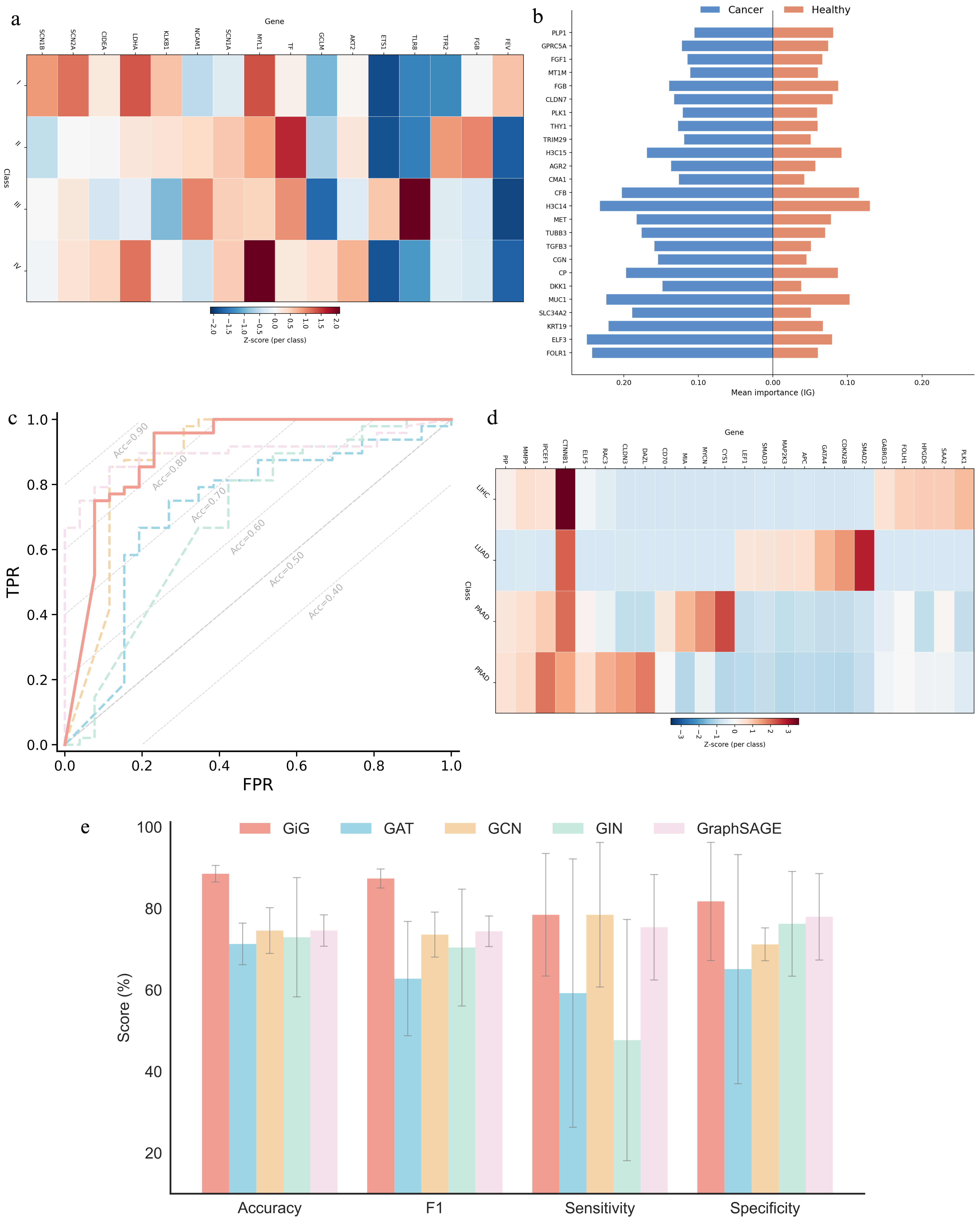}
\caption{\textbf{GiG improves cancer detection, prediction of cancer stages and subtypes and identifies class-dependent cfRNA signals in the RARE-Seq cohort.}
\textbf{a.} Stage-wise z-scored expression heatmap for representative genes. 
\textbf{b.} Integrated Gradients (IG) attribution butterfly plot for  cancer detection. 
\textbf{c.} Macro-averaged ROC curves comparing GiG with GAT, GCN, GIN, and GraphSAGE methods. 
\textbf{d.} Subtype-wise z-scored expression heatmap for representative genes. 
\textbf{e.} Cross-validation performance across accuracy, macro-F1, sensitivity, and specificity. Bars indicate mean scores and error bars indicate variability across folds.}
\label{fig:rareseq_fig1}
\end{figure}

In Figure ~\ref{fig:rareseq_fig1}e, our results demonstrate that GiG consistently outperforms node-level graph neural network baselines - including GAT, GCN, GIN, and GraphSAGE - across binary, stage, and subtype classification tasks (Fig.~\ref{fig:rareseq_fig1}e, \ref{fig:Rareseq_all_task_baselines}). Its performance superiority is most pronounced in the binary classification task, where GiG achieves the highest scores across all evaluation metrics. Specifically, GiG attains an accuracy of 89\% and a macro-F1 score of 88\%, outperforming the strongest baseline by over 15\% in accuracy and 14\% in macro-F1 score. GiG also demonstrates superior sensitivity (88\%) and specificity (89\%), indicating strong and balanced performance across both classes. In contrast, even the strongest baseline model, GCN, achieves only approximately 74\% accuracy and macro-F1, with sensitivity and specificity around 78\%. Overall, this indicates GiG has a more balanced ability to capture minority class signals under significant class imbalance, while node-based models are more prone to favoring dominant classes, a common challenge in cfRNA-based classification tasks. As such, incorporating pathway-structured, sample-specific graph representations in our model enhanced its ability to retrieve weak but biologically meaningful signals embedded within noisy and heterogeneous cfRNA measurements. Together, these results demonstrate that GiG not only improves overall predictive accuracy but also achieves more robust and balanced performance across evaluation metrics, which is particularly critical in clinical settings where both false positives and false negatives carry significant consequences.

Beyond classification metrics, we evaluated GiG's performance using Macro-averaged ROC curves. As shown in Figure ~\ref{fig:rareseq_fig1}c, GiG consistently outperforms node-level baselines across operating thresholds, with the largest separation observed at low false positive rates - a region of particular importance for high-precision classification tasks that are clinically relevant. In this crucial region, GiG achieves higher true positive rates than all baselines, indicating improved sensitivity under stringent false positive constraints. In contrast, node-level models have more gradual increases in the true positive rate and require substantially higher false positive rates to achieve comparable sensitivity.

To characterize the distribution of predictive signals across genes, we analyzed class-wise expression patterns for stage and subtype (Fig.~\ref{fig:rareseq_fig1}a, d). Z-scored profiles reveal that each class exhibits distinct expression patterns across subsets of genes. For example, in Figure ~\ref{fig:rareseq_fig1}a, Stage I samples show elevated expression of SCN1B and reduced expression of TLR8, whereas Stage III demonstrates the opposite pattern, with lower SCN1B expression and higher TLR8 levels. These contrasting patterns indicate that disease stage identity is defined by distinct multi-gene expression signatures rather than the reliance on a single dominant marker. Similarly, Figure ~\ref{fig:rareseq_fig1}d shows distinct expression patterns across cancer subtypes, with each subtype characterized by a unique combination of up- and down-regulated genes. Notably, these differences do not reflect uniform shifts across all features but instead arise from heterogeneous changes across gene subsets. This suggests that discriminative signal in cfRNA data is distributed in a class-dependent manner, with distinct transcriptional patterns underlying different cancer subtypes.

Lastly, we also examined Integrated Gradients (IG) attribution scores using a butterfly plot to assess how predictive signals are allocated across classes in the binary task (Fig.~\ref{fig:rareseq_fig1}b). The results reveal clear class-specific asymmetry in feature importance, with distinct sets of genes contributing preferentially to cancer and healthy predictions. Genes such as FOLR1, ELF3, and KRT19 exhibit elevated attribution for cancer classification but relatively low importance for healthy classification, while a different subset of genes shows higher attribution for healthy samples. The elevated cancer-specific attribution of FOLR1, ELF3, and KRT19 is biologically consistent with their established roles as epithelial and tumor-associated markers. FOLR1 has been linked to tumor and circulating protein levels in ovarian cancer \cite{bax2023folate,leung2013folate}, ELF3 regulates epithelial transcriptional programs involved in tumor progression \cite{ju2024function}, and KRT19 is a canonical epithelial cytokeratin frequently used in circulating tumor cell detection \cite{saloustros2011cytokeratin}. These associations suggest that GiG prioritizes biologically plausible tumor-derived cfRNA signals for cancer classification rather than relying solely on nonspecific plasma RNA variation.

Taken together, these results demonstrate that incorporating pathway-structured representations allowed GiG to consistently improve predictive performance while capturing biologically meaningful and class-specific signal in cfRNA data. The observed gains in accuracy, sensitivity, and macro-F1, along with improved discrimination across ROC thresholds, indicate that pathway-structured representations enhance robustness under conditions of low signal-to-noise ratio, substantial heterogeneity, and class imbalance. Importantly, both expression and attribution analyses show that discriminative information is distributed across coordinated gene subsets and selectively prioritized by the model in a class-dependent manner. As such, these results show that the advantages of pathway-informed graph construction extend beyond well-controlled datasets to ultrasensitive liquid biopsy settings, where reliable molecular inference is difficult yet clinically impactful.

\subsection{GiG demonstrates near-perfect performance and robust generalization on prostate cancer diagnosis in a tissue-based transcriptomic dataset}
Next, we evaluated GiG on a tumor tissue-derived prostate cancer transcriptomic data set \cite{elmarakeby2021biologically}, representing a less noisy setting compared to liquid biopsy data based on cfRNA. Unlike RARE-Seq, where tumor-derived signals are sparse and diluted, tissue-based expression profiles capture more direct molecular programs associated with tumor development and progression. In other words, this dataset is characterized by well-defined molecular profiles and recurrent alterations in key signaling pathways that collectively shape prostate cancer biology, like PI3K/AKT, MAPK, and Wnt signaling. Therefore, this setting allows us to assess whether the advantages of pathway-structured representations become even more pronounced in a higher-signal regime, where biological organization is more clearly represented in the data, compared to the more diffuse and noisy signals observed in cfRNA.\cite{elmarakeby2021biologically}

\begin{figure}[p]
\centering
\includegraphics[width=\linewidth,height=\textheight,keepaspectratio]{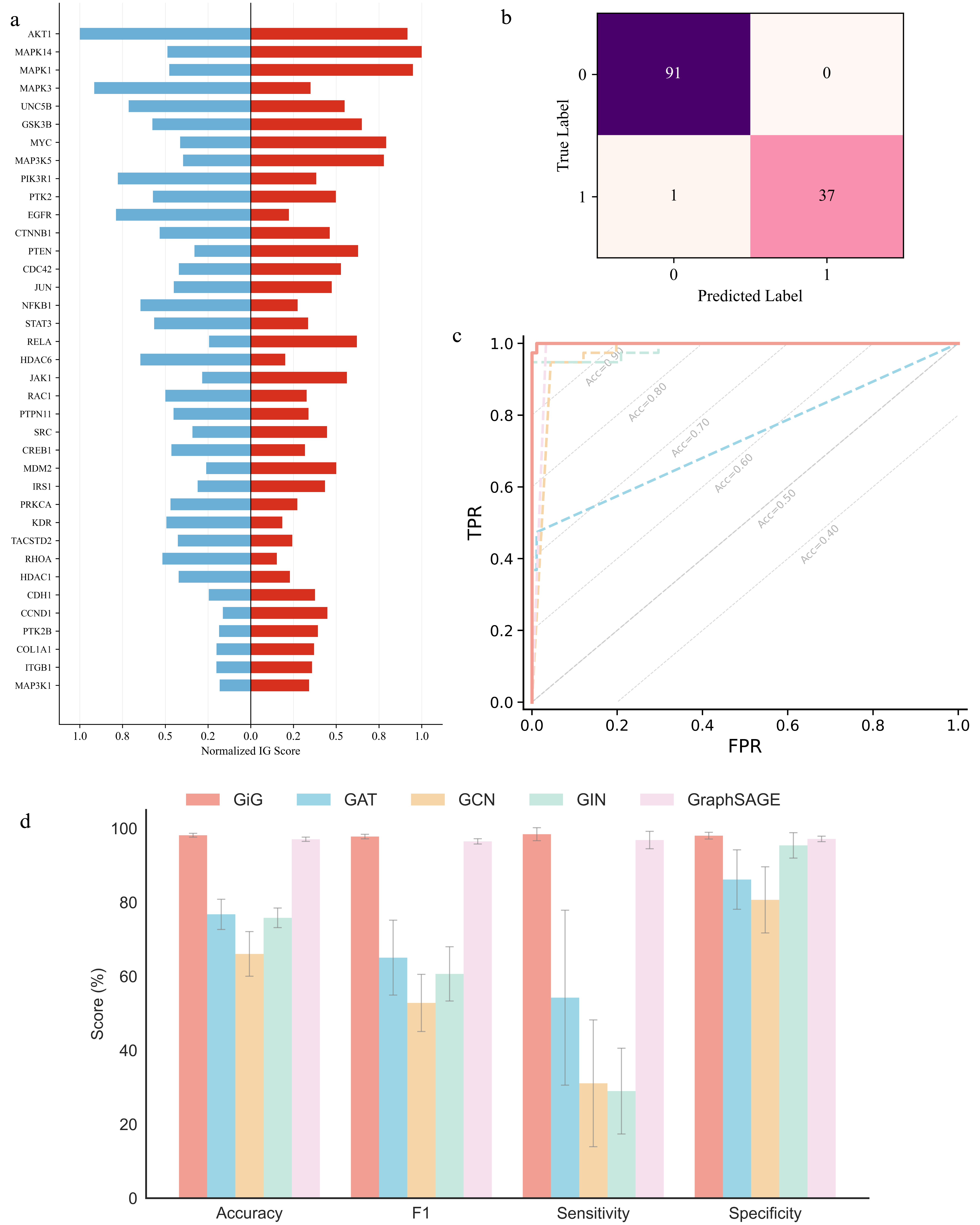}
\caption{\textbf{GiG achieves near-perfect performance in prostate cancer classification.}
\textbf{a.} Integrated Gradients (IG) attribution plot highlighting predictive genes, including canonical pathway components such as AKT1, MAPK1, and CTNNB1. 
\textbf{b.} Confusion matrix for localized versus metastatic prostate cancer classification. 
\textbf{c.} Macro-averaged ROC curves comparing GiG with GAT, GCN, GIN, and GraphSAGE baselines. 
\textbf{d.} Cross-validation performance across accuracy, macro-F1, sensitivity, and specificity. Bars indicate mean scores and error bars indicate variability across folds.}
\label{fig:prostate_cancer_fig}
\end{figure}

As shown in Figure~\ref{fig:prostate_cancer_fig}e, GiG achieves near-perfect performance across all evaluation metrics, substantially outperforming node-level graph neural network baselines. GiG attains accuracy, macro-F1, and sensitivity values approaching 98\%, with consistently low variance across folds. In contrast, node-level baselines - including GAT, GCN, and GIN - exhibit significantly lower and more variable performance, particularly in sensitivity, where several models fall below 60\%. While GraphSAGE achieves competitive accuracy and specificity, GiG still performs better across all four metrics measured. These results indicate that GiG not only improves predictive accuracy but also maintains consistently high recall of the positive class, which is critical for identifying clinically relevant disease states.

Confusion matrix analysis further supports these findings, showing that GiG achieves near-perfect classification with minimal misclassification between localized and metastatic samples (Fig.~\ref{fig:prostate_cancer_fig}b). Evidently, GiG is able to effectively capture discriminative patterns associated with disease progression. Consistent with this observation, macro-averaged ROC curves also demonstrate that GiG outperforms baseline models, achieving near-optimal true positive rates at extremely low false positive rates across operating thresholds (Fig.~\ref{fig:prostate_cancer_fig}d). This indicates strong separability in the learned representations and highlights GiG’s ability to operate effectively under stringent, clinically relevant classification criteria.

Lastly, to investigate the molecular features driving these predictions, we also analyzed Integrated Gradients (IG) attribution scores (Fig.~\ref{fig:prostate_cancer_fig}a). The results reveal that predictive importance is distributed across a set of biologically relevant genes, including AKT1, MAPK1, and CTNNB1. These genes are central components of canonical signaling pathways - such as PI3K/AKT, MAPK, and Wnt signaling - that are known to play critical roles in prostate cancer developments \cite{elmarakeby2021biologically, robinson2015integrative, abida2019genomic, roos2016dna}. Such alignment with established prostate cancer biology suggests that GiG is capable of prioritizing features consistent with underlying disease mechanisms.

Overall, these results show that GiG prioritizes biologically meaningful genes associated with canonical prostate cancer pathways and achieves near-ceiling performance across accuracy, sensitivity, macro-F1, and ROC-based discrimination. Consistent with observations from the RARE-Seq dataset, pathway incorporation in GiG enables highly separable and reliable classification. In other words, GiG is not only robust in noisy, low-signal environments but achieves even stronger performance in higher-signal regimes, where biological organization is more clearly represented and can be more effectively leveraged for clinically useful prediction.

\subsection{GiG achieves scalable multi-class cancer classification across diverse pan-cancer types}
Next, we evaluated GiG on a pan-cancer transcriptomic dataset derived from The Cancer Genome Atlas (TCGA), representing a large-scale, multi-class classification setting with substantial molecular and tissue-level heterogeneity\cite{weinstein2013cancer}. Unlike the binary and stage-specific tasks considered previously, this dataset requires the model to distinguish a diverse set of cancer types, each characterized by different transcriptional programs with varying degrees of similarity. As such, this dataset provides a stringent test of whether GiG can scale effectively and maintain performance in the presence of increased class complexity and inter-class variability.

\begin{figure}[pth]
\centering
\includegraphics[width=\linewidth]{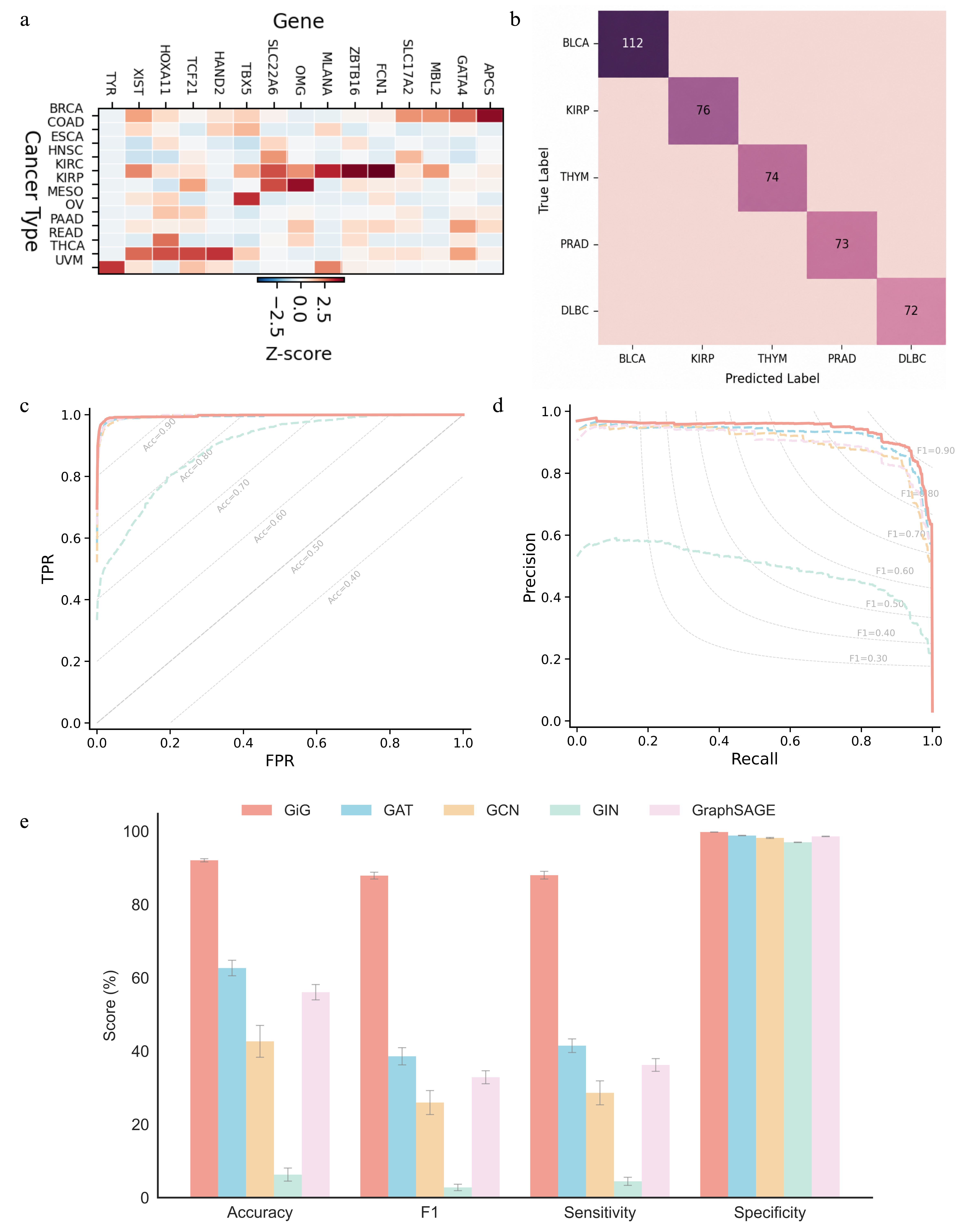}
\caption{\textbf{GiG achieves scalable multi-class classification across pan-cancer types.} \textbf{a.} Z-scored Integrated Gradients attribution heatmap shows class-specific gene importance patterns across cancer types. \textbf{b.} Confusion matrix for the top five cancer classes demonstrates strong diagonal dominance and limited inter-class misclassification. \textbf{c.} Macro-averaged ROC curves comparing GiG with node-level GNN baselines. \textbf{d.} Macro-averaged precision-recall curves showing improved precision across recall thresholds. \textbf{e.} Cross-validation performance across accuracy, macro-F1, sensitivity, and specificity. Bars indicate mean scores and error bars indicate variability across folds.}
\label{fig:pancancer_fig}
\end{figure}

As shown in Figure~\ref{fig:pancancer_fig}e, GiG achieves over 90\% accuracy and steadily outperforms node-level graph neural network baselines across all evaluation metrics. Consistent with observations from both the RARE-Seq and prostate cancer datasets, GiG maintains substantially higher macro-F1 and sensitivity scores than GAT, GCN, GIN, and GraphSAGE, which demonstrates its superior ability to maintain balanced predictions across diverse cancer classes. Although specificity remains relatively similar across models, GiG still achieves the strongest overall performance. Together, these results show that GiG is able to provide robust and accurate classification even in a large-scale, multi-class setting characterized by substantial biological heterogeneity.

Confusion matrix analysis further illustrates GiG’s ability to distinguish between cancer types with high fidelity, showing strong diagonal dominance and minimal confusion across classes (Fig.~\ref{fig:pancancer_fig}b). This suggests that the model effectively captures discriminative features unique to each cancer type, despite the biological overlap between certain classes. Consistent with these observations, macro-averaged ROC curves demonstrate that GiG outperforms baseline models across operating thresholds, maintaining superior true positive rates at low false positive rates even in this multi-class setting (Fig.~\ref{fig:pancancer_fig}c). Similarly, precision-recall curves show that GiG achieves consistently higher precision across a wide range of recall values compared to baseline models (Fig.~\ref{fig:pancancer_fig}d), indicating improved performance under class imbalance, reinforcing its ability to make reliable predictions across diverse cancer types.

To further examine the molecular features driving pan-cancer classification, we analyzed Integrated Gradients (IG) attribution scores across cancer types using a heatmap representation (Fig.~\ref{fig:pancancer_fig}a). As shown in Figure 4a, each cancer type exhibits distinct attribution patterns across different subsets of genes. Evidently, GiG distributes predictive importance in a class-dependent manner, rather than relying on a shared set of dominant features across all cancer types. These results suggest that pathway-informed graph representations enable GiG to capture discriminative molecular signals specific to individual cancer types despite extensive transcriptomic heterogeneity. Furthermore, several highly attributed genes are biologically consistent with known lineage-specific and cancer-associated literature. For example, APCS exhibits elevated attribution in liver hepatocellular carcinoma (LIHC), consistent with its established role as a liver-associated acute phase protein and biomarker in hepatic disease. Similarly, ZBTB16 and MLLT6 show elevated importance in kidney cancers (KIRC/KIRP), while TYR demonstrates strong attribution in uveal melanoma (UVM), consistent with its role in melanocyte differentiation and melanin biosynthesis. GATA4 also exhibits subtype-specific attribution patterns, aligning with its known involvement in lineage specification and tumor-associated transcriptional regulation across multiple cancer contexts.

Together, these results demonstrate that GiG effectively scales to complex multi-class classification tasks, maintaining strong performance, accurate and balanced predictions across diverse cancer types. GiG's ability to capture class-specific gene expression patterns and achieve superior accuracy and precision indicates that pathway-informed graph representations provide a robust framework for modeling large-scale, heterogeneous transcriptomic datasets. Importantly, these findings suggest that the advantages of GiG extend beyond binary classification task shown in RARE-Seq, and low-class classification task in Prostate Cancer dataset; rather it extends to high-dimensional, multi-class problems demonstrated in this pan-cancer dataset, where capturing both shared and distinct biological structure is essential for accurate and clinically meaningful prediction.

\section{Discussion}

In this study, we introduced Graph-in-Graph (GiG), a patient-specific graph learning framework that integrates transcriptomic measurements with curated biological pathway topology. Across liquid biopsy, tissue-based prostate cancer, and pan-cancer classification tasks, GiG consistently improved predictive performance relative to standard graph neural network baselines. These results suggest that representing each patient as an individualized pathway-structured graph can improve clinical prediction from transcriptomic data, particularly when discriminative signals are distributed across multiple genes and pathway modules rather than concentrated in a small number of dominant markers.

A central question is whether pathway topology contributes structure beyond what node features already encode. We replaced the WikiPathways-derived edges with two random graph controls while keeping every other component of the model identical. Erd\H{o}s--R\'enyi rewiring preserves only the edge count. Degree-preserving rewiring keeps the degree of each node but randomises the connectivity elsewhere. The performance loss under each control measures how much the classifier depends on pathway topology for each task. Graphlet orbit statistics computed on real pathway graphs and both random controls confirm that the curated topology carries closed-motif structure that neither random family reproduces (Fig.~\ref{fig:rareseq_orbit_distributions}, Fig.~\ref{fig:pancancer_orbit_distributions}, Fig.~\ref{fig:prostate_cancer_orbit_distributions}).

On RareSeq binary cancer detection, the real pathway graph reached \textbf{89.2\%} accuracy and \textbf{87.7\%} macro-F1. Erd\H{o}s--R\'enyi rewiring reduced performance to \textbf{73.0\%} accuracy and \textbf{72.5\%} macro-F1. Degree-preserving rewiring reduced performance further to \textbf{67.6\%} accuracy and \textbf{56.0\%} macro-F1. Disrupting pathway topology cost between \textbf{16.2\%} and \textbf{21.6\%} in accuracy and between \textbf{15.2\%} and \textbf{31.7\%} in macro-F1. Neither the edge count nor the degree sequence alone recovered the pathway signal. In the low-signal cfRNA setting, pathway structure organizes weak patient-specific molecular signals into functional neighborhoods where coordinated dysregulation aggregates more effectively than across random connectivity.

The subtype and stage tasks reveal the same pattern through macro-F1 rather than accuracy. For subtype classification, the real graph reached \textbf{93.8\%} accuracy and \textbf{54.2\%} macro-F1, while both random controls held at \textbf{87.5\%} accuracy but collapsed to \textbf{23.3\%} macro-F1. For stage classification, the real graph reached \textbf{83.3\%} accuracy and \textbf{52.0\%} macro-F1, while both controls reached \textbf{77.1\%} accuracy and \textbf{21.8\%} macro-F1. The accuracy drop of \textbf{6.3\%} in both tasks understates the loss. The control models default to majority-class prediction, and the \textbf{87.5\%} subtype control accuracy matches the LUAD prevalence in the cohort to the nearest decimal. Macro-F1 collapses by \textbf{30.8\%} on subtype and \textbf{30.2\%} on stage. Pathway topology contributes most where the model must separate clinically meaningful subclasses rather than recover the dominant class.

The pan-cancer ablation extends this result across 32 cancer types. The real pathway graph reached \textbf{92.4\%} accuracy and \textbf{88.9\%} macro-F1. Erd\H{o}s--R\'enyi rewiring reduced accuracy by \textbf{24.2\%} and macro-F1 by \textbf{31.8\%}. Degree-preserving rewiring reduced accuracy by \textbf{15.3\%} and macro-F1 by \textbf{18.7\%}. Random topologies retain nontrivial predictive ability, confirming that expression-derived node features carry disease-discriminative information on their own. Neither control approached the real-graph result. Across 32 cancer classes with substantial transcriptional overlap, curated pathway topology provides an inductive bias that the model cannot recover from node features alone.

In contrast, the prostate cancer ablation showed only modest dependence on exact pathway topology. Relative to the real pathway graph, Erdős–Rényi random wiring reduced accuracy by only \textbf{1.55 percentage points} and macro-F1 by \textbf{1.87 percentage points}, while degree-preserving rewiring reduced accuracy by only \textbf{0.78 percentage points} and macro-F1 by \textbf{0.94 percentage points}. Such small decreases suggest that, in the prostate cancer cohort, transcriptomic node features already contained highly separable disease signal, making the model less sensitive to the precise arrangement of pathway edges. This result is useful because it prevents overclaiming: GiG does not require that pathway topology dominate in every dataset. Rather, the relative contribution of topology depends on the strength of the molecular signal and the difficulty of the classification task.

Together, these experiments separate the contribution of molecular signal from the contribution of biological structure. Across all datasets, random graphs retained nontrivial predictive ability, confirming that patient-specific expression features remain the primary source of discriminative information. However, the magnitude of performance loss after topology disruption shows that curated pathway structure contributes meaningful organization, especially in PanCancer and RARE-Seq, where biological heterogeneity and class imbalance make the learning problem more difficult. The strongest interpretation is therefore not that graph topology alone predicts cancer state, but that biologically meaningful topology improves how molecular measurements are represented, propagated, and aggregated.

A careful way to state the conclusion is that pathway-based graphs improve prediction by structuring patient-specific molecular measurements according to known biological relationships. The ablation results do not imply that every pathway edge is correct or equally informative, and they do not imply that topology is always the dominant signal. In the prostate cancer task, node features appear to dominate; in RARE-Seq and PanCancer, pathway topology provides a clearer performance advantage. This context-dependent pattern is consistent with the intended role of the Graph-in-Graph framework: to combine data-driven molecular measurements with biological priors, so that disease-relevant transcriptomic perturbations can be learned in pathway context rather than as isolated tabular features.

Overall, the ablation experiments support the central hypothesis that biologically grounded pathway topology contributes to GiG performance, but they also show that its contribution varies with the structure of the prediction task. The largest gains from real pathway topology were observed in the settings where biological signal is most difficult to decode, including low-signal cfRNA classification and heterogeneous pan-cancer classification. These results suggest that patient-specific pathway graphs can serve as a useful inductive bias for clinical transcriptomic modeling, improving robustness and interpretability while preserving the molecular organization of disease biology. As larger and more diverse clinical omics datasets become available, GiG provides a flexible framework for integrating curated biological knowledge directly into patient-level predictive models.

\section{Methods}
Graph-in-Graph (GiG) was designed to convert transcriptomic measurements into patient-specific molecular interaction graphs that could be used directly for graph-level learning and predictions. The graph construction pipeline consisted of four main stages: gene identifier canonicalization, sample-specific selection of transcriptionally dysregulated genes, mapping of selected genes to curated biological pathways, and assembly of a patient-level graph by merging the corresponding pathway interaction graphs. The resulting graph for each patient encoded both sample-specific molecular activity and prior biological pathway structure. These graphs were then used as inputs to graph neural network classifiers for downstream disease classification tasks.

\subsection{Transcriptomic preprocessing and gene identifier canonicalization}

For each dataset, raw expression matrices were first reformatted so that rows corresponded to genes and columns corresponded to samples during graph construction. Gene identifiers were canonicalized to HGNC gene symbols to ensure consistent alignment between measured transcriptomic expression matrix and WikiPathways-derived pathway topology. For datasets with Ensembl gene identifiers, version suffixes were removed before identifiers were queried using MyGeneInfo, and only genes annotated as protein-coding with valid HGNC symbols were retained. Genes without valid HGNC mappings were excluded from downstream graph construction, ensuring that every graph node corresponded to a uniquely resolved, measured gene feature, reducing the risk of introducing unmatched or biologically ambiguous nodes into the patient-specific pathway graphs.

To mitigate the effect of low-variance and less informative features, highly variable genes were selected before graph construction. Specifically, genes were ranked by variance across samples, and the most highly variable genes were retained for subsequent pathway mapping. In the implemented graph-construction pipeline, the RARE-Seq and prostate cancer cohorts were filtered to the top 10,000 highly variable genes before sample-specific graph assembly, whereas the pan-cancer dataset used the pre-aligned gene universe provided as input. The pan-cancer cohort was not subjected to additional HVG selection because the input matrix was already provided as a pre-aligned gene expression matrix for cross-cancer comparison, and introducing an additional filtering step could remove subtype specific genes with lower global variance but potential relevance to individual cancer classes. After preprocessing, the resulting expression matrix was denoted as
\[
X \in \mathbb{R}^{G \times N},
\]
where \(G\) is the number of retained genes and \(N\) is the number of samples.

\subsection{Sample-specific selection of dysregulated genes}

For each sample, genes were selected on the basis of their relative expression deviation within that sample. Let \(x_{gi}\) denote the expression value of gene \(g\) in sample \(i\). Expression values were standardized independently within each sample across all retained genes:
\[
z_{gi} = \frac{x_{gi} - \mu_i}{\sigma_i},
\]
where \(\mu_i\) and \(\sigma_i\) are the mean and standard deviation of expression values across all genes in sample \(i\). The absolute z-score,
\[
m_{gi} = |z_{gi}|,
\]
was used as a direction-independent measure of transcriptional dysregulation, allowing both highly up-regulated and highly down-regulated genes to contribute to graph construction.

For each sample \(i\), a dysregulated gene set \(D_i\) was defined by retaining genes whose absolute z-score exceeded a dataset-specific percentile threshold:
\[
D_i = \{g \in G : |z_{gi}| \geq Q_{\tau}(|z_i|)\},
\]
where \(Q_{\tau}(|z_i|)\) is the \(\tau\)-th percentile of absolute z-score values across genes in sample \(i\). For the primary GiG pipeline, patient-specific graphs were constructed using an 80th percentile threshold on the absolute z-score distribution. This threshold retained the most dysregulated genes in each sample while preserving sufficient pathway coverage for graph construction. Alternative percentile thresholds, including 10th, 20th, and 90th percentile cutoffs, were also evaluated during pipeline development: the 80th percentile threshold was selected for the final analyses because it provided a practical balance between sample-specific sparsity and biological pathway representation.

\subsection{Mapping dysregulated genes to WikiPathways}

Selected genes were mapped to curated biological pathways using WikiPathways \cite{agrawal2024wikipathways}. For each gene \(g \in D_i\), pathway membership was queried in WikiPathways using the human gene identifier namespace. The returned pathway identifiers were represented as WikiPathways IDs (WPIDs). For sample \(i\), the set of selected pathways was defined as
\[
P_i = \bigcup_{g \in D_i} P(g),
\]
where \(P(g)\) denotes the set of curated pathways containing gene \(g\) in the WikiPathways database. Genes without associated pathways were recorded and excluded from pathway selection. Gene-to-pathway mappings were cached locally to avoid repeated queries and to ensure that identical gene inputs yielded the same pathway set across experiments. Overall, this procedure allowed each sample to select a distinct set of pathways based on its unique transcriptional profile. As a result, the final graph topology was not fixed across all samples, but instead reflected the pathways associated with the most dysregulated genes in each individual sample.

\subsection{Construction of pathway-level interaction graphs}

For each WPID selected for a sample, the corresponding pathway was retrieved from WikiPathways in Graphical Pathway Markup Language (GPML) format and cached locally for reuse. Each GPML file was parsed to extract two classes of pathway elements: molecular data nodes and interaction elements. Data nodes representing genes, RNA molecules, and proteins were retained as candidate graph nodes, whereas non-HGNC-resolvable elements - including microRNAs-specific nodes - were excluded. Retained molecular nodes were initially extracted using their GPML pathway labels and available cross-reference identifiers. Before sample-level graph assembly, these nodes were canonicalized to HGNC gene symbols using available Ensembl, UniProt, or Entrez Gene cross-references queried through the \texttt{mygene} \cite{wu2013biogps} Python package. Overall, this workflow ensured that each final graph node corresponded to a measured gene feature in the input transcriptomic expression matrix.

Next, edges were inferred from GPML interaction elements and added to the pathway graph. Each interaction was inspected for graphical points that referenced retained molecular nodes through GPML \texttt{GraphRef} attributes. Interactions between two retained molecular nodes were represented as undirected edges connecting the corresponding nodes. When an interaction referenced more than two retained molecular nodes, all pairwise edges among those nodes were added. The resulting HGNC-standardized pathway \(p\) was therefore represented as an undirected molecular interaction graph:
\begin{equation}
H_p = (V_p, E_p)
\label{eq:pathway_graph}
\end{equation}

Here, \(V_p\) denotes the set of retained molecular nodes in pathway \(p\), and \(E_p\) denotes the set of curated pathway interactions extracted from the GPML representation. Node labels in \(H_p\) were then canonicalized to HGNC gene symbols. Specifically, available Ensembl, UniProt, and Entrez Gene cross-references were queried using the \texttt{mygene} Python client for MyGene.info, and nodes that could not be mapped to valid HGNC symbols were removed. After canonicalization, pathway graphs were again filtered to retain only genes present in the processed mRNA expression matrix. Self-loops introduced by identifier collapsing were removed, and the resulting HGNC-standardized pathway graph was denoted as follows:
\begin{equation}
\tilde{H}_p = (\tilde{V}_p, \tilde{E}_p), \qquad \tilde{V}_p \subseteq G
\label{eq:hgnc_pathway_graph}
\end{equation}
where \(G\) denotes the set of measured genes retained in the processed transcriptomic expression matrix. Note that each processed pathway graph was cached by WPID and reused across samples, ensuring that pathway retrieval, GPML parsing, and HGNC canonicalization were performed once per pathway rather than repeatedly for every sample. After individual pathway graphs were constructed, all pathways selected for a given sample were merged into a single undirected molecular interaction graph for that patient. Overall, this procedure preserved WikiPathways-derived gene-gene connectivity while focusing the final graph on genes and pathways associated with the strongest patient-specific expression deviations.

\subsection{Assembly of patient-specific pathway graphs}

For each sample \(i\), all processed pathway graphs corresponding to the selected pathway set \(P_i\) were merged into a single patient-specific graph. The resulting graph was denoted as follows:
\begin{equation}
\mathcal{G}_i = (V_i, E_i) = \bigcup_{p \in P_i} \tilde{H}_p
\label{eq:patient_graph}
\end{equation}

Here, \(\mathcal{G}_i\) denotes the patient-specific graph for sample \(i\), \(P_i\) denotes the set of curated biological pathways selected from WikiPathways for that sample, and \(\tilde{H}_p\) denotes the HGNC-standardized \cite{seal2026genenames} graph for pathway \(p\). The node and edge sets of \(\mathcal{G}_i\) were defined as the union of the node and edge sets from all selected pathway graphs:
\begin{equation}
V_i = \bigcup_{p \in P_i} \tilde{V}_p,
\qquad
E_i = \bigcup_{p \in P_i} \tilde{E}_p
\label{eq:patient_graph_union}
\end{equation}

Here, \(V_i\) and \(E_i\) denote the final node and edge sets of sample \(i\), whereas \(\tilde{V}_p\) and \(\tilde{E}_p\) denote the HGNC-standardized \cite{seal2026genenames} node and edge sets of individual pathway \(p\). Genes appearing in multiple pathways were collapsed into a single node based on their HGNC symbol, while edges from all selected pathways were retained. This construction allowed shared genes to form bridges between pathway modules, producing an integrated sample-specific molecular interaction network. Self-loops were removed after merging, and each final graph was then used as the patient-level input for downstream graph neural network modeling. In this representation, nodes correspond to measured HGNC genes, edges correspond to curated WikiPathways interactions, and graph topology varies across patients according to the pathways selected from each patient’s gene expression profile.

\subsection{Graph-level dataset assembly}

For each prediction task, patient-specific pathway graphs were paired with matched transcriptomic expression profiles and task labels before model training. Upon parsing the graph files, only samples with corresponding expression profile, clinical label, and patient-specific graph were retained. After sample alignment, task labels were integer-encoded and each patient graph was converted into a PyTorch Geometric \texttt{Data} object for graph-level classification.

For a sample \(i\), the graph object was represented as:
\begin{equation}
\mathcal{D}_i =
\left(
X_i,
\mathbf{g}_i,
\mathrm{edge\_index}_i,
y_i
\right).
\label{eq:pyg_data_object}
\end{equation}

Here, \(X_i \in \mathbb{R}^{|V_i| \times 2}\) denotes the two-channel node feature matrix for graph \(\mathcal{G}_i\), \(\mathbf{g}_i\) denotes the integer-encoded gene identity for each node, \(\mathrm{edge\_index}_i\) denotes the graph connectivity used for message passing, and \(y_i\) denotes the encoded sample-level class label. Graphs with fewer than two nodes or without valid edges were excluded from model training. Because the pathway graphs were treated as undirected for message passing, each edge was represented in both directions in the PyTorch Geometric edge index.

\subsection{Node feature construction}

Each graph node represented a gene and was assigned two scalar features. The first feature captured the patient-specific expression level of that gene after within-sample normalization. For sample \(i\), expression values were standardized across genes so that each node feature reflected the relative expression of gene \(g\) within that patient:
\begin{equation}
z_{gi} =
\frac{T(x_{gi}) - \mu_i}{\sigma_i + \epsilon}.
\label{eq:model_samplewise_zscore}
\end{equation}

Here, \(x_{gi}\) is the expression value of gene \(g\) in sample \(i\), \(T(\cdot)\) denotes the transformation applied before normalization, \(\mu_i\) and \(\sigma_i\) are the mean and standard deviation across genes within sample \(i\), and \(\epsilon\) is a small numerical constant used to avoid division by zero. When appropriate for the input scale, expression values were transformed using \(\log(1+x)\) before standardization.

The second node feature summarized the cohort-level co-expression context of each gene. Expression values were first standardized gene-wise across samples:
\begin{equation}
\tilde{x}_{gi} =
\frac{T(x_{gi}) - \mu_g}{\sigma_g + \epsilon},
\label{eq:model_genewise_zscore}
\end{equation}
where \(\mu_g\) and \(\sigma_g\) are the mean and standard deviation of gene \(g\) across samples. Pairwise gene correlation was then computed as:
\begin{equation}
r_{gh} =
\frac{1}{N}
\sum_{i=1}^{N}
\tilde{x}_{gi}\tilde{x}_{hi}.
\label{eq:pcc_correlation}
\end{equation}

For each gene \(g\), the final Pearson correlation coefficient (PCC) node feature was defined as the mean absolute correlation with its top \(k\) most correlated genes, excluding self-correlation:
\begin{equation}
c_g =
\frac{1}{k}
\sum_{h \in \mathrm{TopK}(g)}
|r_{gh}|,
\qquad k=50.
\label{eq:pcc_node_feature}
\end{equation}

The resulting feature vector for gene \(g\) in sample \(i\) was:
\begin{equation}
\mathbf{x}_{gi} =
\begin{bmatrix}
z_{gi} \\
c_g
\end{bmatrix}.
\label{eq:node_feature_vector}
\end{equation}

Thus, each node carried both a patient-specific expression feature and a cohort-level co-expression feature, while the graph structure preserved curated pathway topology.

\subsection{Gene identity encoding}

In addition to the two scalar node features, each gene was assigned an integer index from a gene vocabulary constructed from the processed expression matrix. This index was passed through a learnable embedding layer during model training. For a node corresponding to gene \(g\), the gene identity embedding was defined as:
\begin{equation}
\mathbf{e}_g =
\mathbf{E}[\mathrm{id}(g)],
\label{eq:gene_embedding}
\end{equation}
where \(\mathbf{E}\) is a trainable embedding matrix and \(\mathrm{id}(g)\) is the integer index assigned to gene \(g\). Genes not found in the expression-derived vocabulary were assigned to a reserved unknown index. This embedding allowed the model to preserve molecular identity across patient-specific graphs whose node sets and connectivity patterns varied across samples.

The two-channel expression feature vector was projected into the same latent dimension as the gene identity embedding:
\begin{equation}
\mathbf{q}_{gi} =
\mathrm{ReLU}
\left(
W_x \mathbf{x}_{gi} + \mathbf{b}_x
\right).
\label{eq:expression_projection}
\end{equation}

The initial node representation was then constructed by concatenating the projected expression features with the learnable gene identity embedding and passing the result through a node-level multilayer perceptron:
\begin{equation}
\mathbf{h}_{gi}^{(0)}
=
f_{\mathrm{node}}
\left(
\mathbf{q}_{gi} \, \Vert \, \mathbf{e}_g
\right),
\label{eq:initial_node_embedding}
\end{equation}
where \(\Vert\) denotes concatenation. In all experiments, the gene identity embedding dimension was 64, the expression projection dimension was 64, and the hidden dimension after the node MLP was 128.

\subsection{Graph neural network architecture}

GiG was implemented as a graph-level classifier operating on patient-specific pathway graphs. After construction of the initial node embeddings, two graph neural network message-passing layers were applied. The general update rule was:
\begin{equation}
\mathbf{H}_i^{(\ell+1)}
=
\mathrm{GNN}_{\theta}^{(\ell)}
\left(
\mathbf{H}_i^{(\ell)},
\mathrm{edge\_index}_i
\right),
\qquad \ell \in \{0,1\}.
\label{eq:gnn_update}
\end{equation}

Here, \(\mathbf{H}_i^{(\ell)}\) is the matrix of node embeddings for sample \(i\) at layer \(\ell\), and \(\mathrm{GNN}_{\theta}^{(\ell)}\) denotes a backbone-specific message-passing layer. Four GNNs were evaluated within the same graph-level framework: graph convolutional networks (GCN), GraphSAGE, graph isomorphism networks (GIN), and GAT. After message passing, node embeddings were aggregated using global mean pooling to obtain a graph-level patient representation:
\begin{equation}
\mathbf{h}_{\mathcal{G}_i}
=
\frac{1}{|V_i|}
\sum_{v \in V_i}
\mathbf{h}_{vi}^{(2)}.
\label{eq:global_mean_pool}
\end{equation}

The pooled graph embedding was passed through a classifier head consisting of a linear layer, ReLU activation, dropout, and a final linear output layer:
\begin{equation}
\hat{\mathbf{y}}_i =
f_{\mathrm{head}}
\left(
\mathbf{h}_{\mathcal{G}_i}
\right).
\label{eq:classifier_head}
\end{equation}

The output dimension of the final layer was set to the number of classes in the corresponding prediction task. This design allowed the same patient-specific graph representation to be evaluated across multiple message-passing backbones while keeping node feature construction, gene identity encoding, pooling strategy, and classifier architecture fixed.

\subsection{Model training and class-weighted optimization}

Models were trained using stratified train-test splits, with 80\% of samples used for training and 20\% held out for evaluation. Splits were generated using a fixed random seed and were stratified by class label when applicable. All backbones for a given task were trained and evaluated using the same split.

Class imbalance was addressed using class-weighted cross-entropy loss. For a task with \(C\) classes, the weight for class \(c\) was computed from the number of training samples \(n_c\) in that class:
\begin{equation}
w_c =
\frac{1/n_c}{\sum_{j=1}^{C} 1/n_j}
C.
\label{eq:class_weight}
\end{equation}

The training objective for a mini-batch \(\mathcal{B}\) was:
\begin{equation}
\mathcal{L}
=
-\frac{1}{|\mathcal{B}|}
\sum_{i \in \mathcal{B}}
w_{y_i}
\log
\left(
\frac{
\exp(\hat{y}_{i,y_i})
}{
\sum_{c=1}^{C} \exp(\hat{y}_{i,c})
}
\right).
\label{eq:weighted_cross_entropy}
\end{equation}

Models were optimized using AdamW with a learning rate of \(10^{-3}\), weight decay of \(10^{-4}\), batch size of 32, and dropout probability of 0.2. Gradients were clipped to a maximum norm of 5.0 during training. Training was performed on CUDA when available. For each backbone, the checkpoint with the best held-out performance was saved for downstream evaluation and interpretation. Model performance was summarized using accuracy and macro-averaged F1 score, with macro-F1 used to account for class imbalance across prediction tasks.

\subsection{Performance Evaluation}

Model predictions were evaluated on held-out test graphs. Accuracy and macro-averaged F1 score were computed for each backbone and task. Accuracy was defined as:
\begin{equation}
\mathrm{Accuracy}
=
\frac{1}{N}
\sum_{i=1}^{N}
\mathbb{I}
\left(
\hat{y}_i = y_i
\right),
\label{eq:accuracy}
\end{equation}
where \(N\) is the number of test samples and \(\mathbb{I}\) is the indicator function. Macro-F1 was computed as the unweighted mean of class-specific F1 scores:
\begin{equation}
\mathrm{F1}
=
\frac{1}{C}
\sum_{c=1}^{C}
\frac{
2\,\mathrm{Precision}_c\,\mathrm{Recall}_c
}{
\mathrm{Precision}_c+\mathrm{Recall}_c
}.
\label{eq:macro_f1}
\end{equation}

For each task, the four GNN backbones were trained independently and summarized in a backbone comparison table. The best-performing checkpoint for each backbone was saved for downstream analysis. For the prostate cancer task, a model bundle was additionally saved containing the trained weights, model hyperparameters, label classes, gene vocabulary, train-test split indices, and node-name metadata. This enabled interpretability analyses to map model-derived attribution scores back to HGNC gene symbols without retraining the model.

For downstream figures, additional metrics including sensitivity, specificity, receiver operating characteristic curves, and precision-recall curves were computed from model predictions or prediction scores using the same held-out evaluation split. Sensitivity and specificity were defined for each class as:
\begin{equation}
\mathrm{Sensitivity}_c =
\frac{\mathrm{TP}_c}
{\mathrm{TP}_c+\mathrm{FN}_c},
\qquad
\mathrm{Specificity}_c =
\frac{\mathrm{TN}_c}
{\mathrm{TN}_c+\mathrm{FP}_c}.
\label{eq:sensitivity_specificity}
\end{equation}

For multiclass tasks, class-wise metrics were macro-averaged unless otherwise specified. This evaluation framework allowed direct comparison of graph neural network backbones under matched data splits, graph inputs, node features, and optimization settings.

\section{Datasets}

Unless otherwise specified, highly variable genes (HVGs) were selected based on variance across samples or cells within each dataset. All tabular inputs were standardized prior to modeling.

\subsection{Circulating cell-free RNA profiling (RARE-Seq).}

We evaluated GiG on the RARE-Seq circulating cell-free RNA (cfRNA) cohort, a multi-cancer liquid biopsy dataset designed to detect low-abundance tumor-derived transcripts in plasma \cite{nesselbush2025ultrasensitive}. This dataset contains plasma cfRNA expression profiles from healthy individuals and patients with multiple cancer types, with available annotations for cancer status, tissue of origin, and clinical stage. Because cfRNA measurements contain tumor-derived signal diluted within a heterogeneous background of circulating RNA, this cohort provides a challenging setting for evaluating whether pathway-structured graph representations can recover weak but biologically meaningful disease signals. The input expression matrix was restricted to Ensembl gene identifiers. Ensembl version suffixes were removed, and identifiers were mapped to HGNC gene symbols using protein-coding annotations. Genes without valid HGNC symbols were excluded. To reduce the contribution of low-variance and noise-dominated features, the top 10,000 highly variable genes were selected based on variance across samples before patient-specific pathway graph construction. Three classification tasks were defined from the same processed expression and graph inputs. For binary cancer detection, samples were grouped as Healthy or Cancer, excluding categories that were not part of either class definition. For subtype classification, cancer samples were mapped to LUAD, LIHC, PAAD, and PRAD categories, with LUAD (TKI) samples grouped under LUAD. For stage classification, only cancer samples with available stage annotations were retained and classified as stage I, II, III, or IV. Patient-specific pathway graphs were then constructed from the most dysregulated genes in each sample and paired with the corresponding task labels for downstream graph-level classification.

\subsection{Prostate cancer transcriptomic dataset.}

We next evaluated GiG on a tissue-derived prostate cancer transcriptomic dataset containing localized and metastatic prostate cancer samples. In contrast to cfRNA profiling, tissue-based transcriptomic measurements capture more direct tumor-associated molecular programs and therefore represent a higher-signal setting for evaluating pathway-informed graph learning. This dataset was used to test whether patient-specific pathway graphs could distinguish clinically meaningful prostate cancer disease states while prioritizing genes involved in established prostate cancer signaling programs, including PI3K/AKT, MAPK, and Wnt-related pathways. The raw prostate cancer expression matrix was reformatted so that samples and genes were consistently aligned with the graph-construction pipeline. Gene symbols were already provided in HUGO/HGNC-compatible format, so no additional gene identifier conversion was required. Genes were ranked by variance across samples, and the top 10,000 highly variable genes were retained before graph construction. Patient-specific pathway graphs were generated using the same WikiPathways-based pipeline applied to the other cohorts. The classification task was formulated as binary prediction of localized versus metastatic prostate cancer. Each sample was retained only when a matched expression profile, class label, and patient-specific pathway graph were available.

\subsection{Pan-cancer transcriptomic dataset.}

To evaluate GiG in a large multiclass setting, we used a pan-cancer transcriptomic dataset derived from The Cancer Genome Atlas (TCGA) \cite{weinstein2013cancer}. This cohort spans diverse tumor types and tissue lineages, providing a stringent benchmark for testing whether pathway-informed graph representations can scale beyond binary classification to a high-dimensional multiclass prediction problem. Unlike the RARE-Seq and prostate cancer tasks, the pan-cancer task requires the model to distinguish many cancer types with partially overlapping transcriptional programs, making balanced classification and class-specific feature attribution particularly important. The pan-cancer input matrix was provided as a pre-aligned gene expression matrix and was used directly as the gene universe for graph construction. No additional HVG filtering was applied at this stage in order to preserve the aligned gene set across cancer types and avoid removing genes with potential lineage- or subtype-specific relevance. For each sample, genes with the strongest within-sample expression deviations were selected and mapped to WikiPathways to construct a patient-specific pathway graph. The task was formulated as 32-class cancer type classification. Each graph was paired with the corresponding cancer type label and used for graph-level model training and evaluation.

\bibliography{main}

\clearpage
\newpage
\appendix
\setcounter{figure}{0}
\renewcommand{\thefigure}{S\arabic{figure}}
\setcounter{table}{0}
\renewcommand{\thetable}{S\arabic{table}}

\section*{Supplementary Materials}
\begin{figure}[pb!]
\centering
\includegraphics[width=\linewidth]{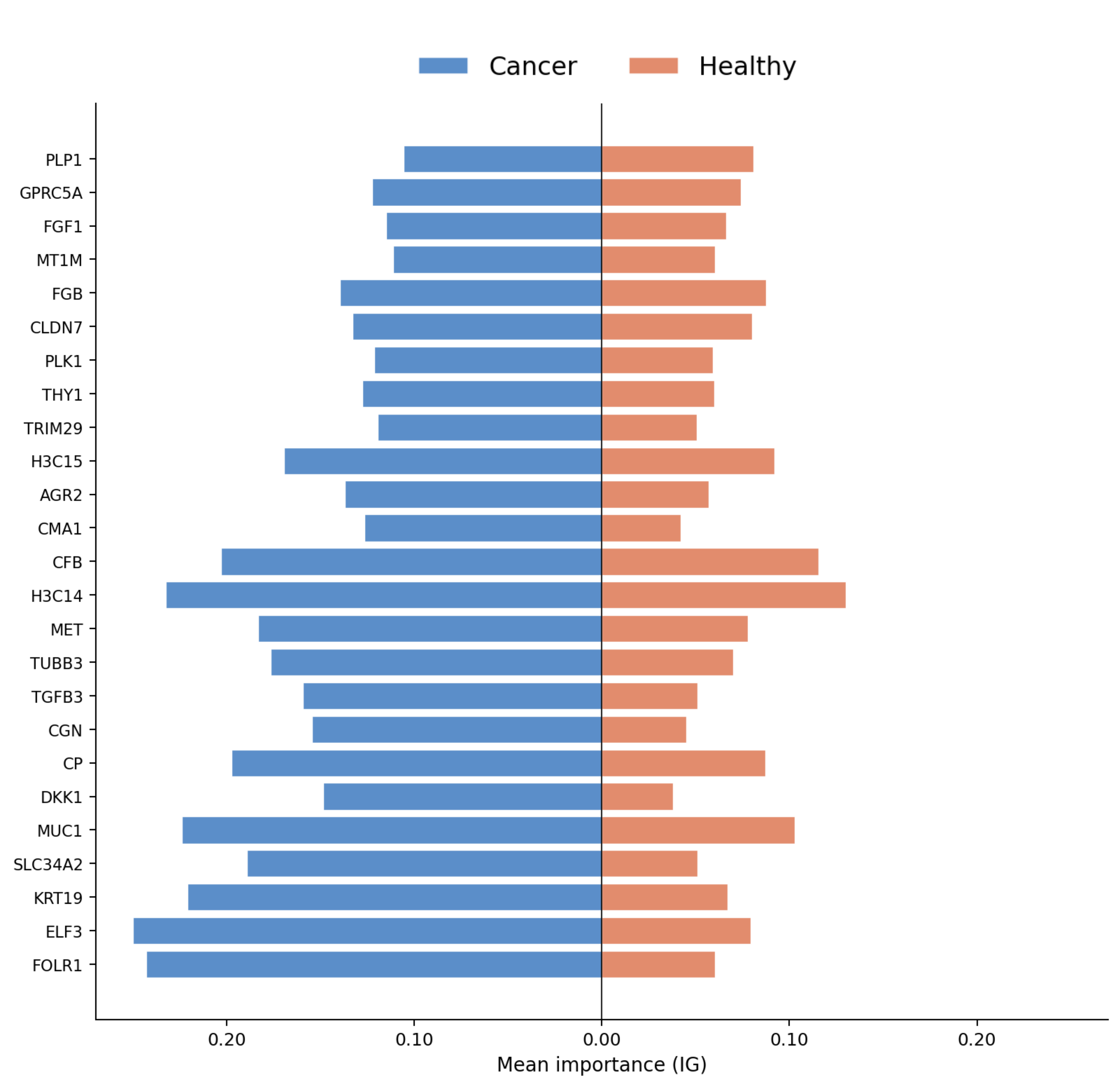}
\caption{\textbf{Integrated Gradients attribution map for RARE-Seq binary classification.}
Class-specific Integrated Gradients (IG) attribution scores for the RARE-Seq cancer-versus-healthy classification task. Each bar represents the mean IG importance of a gene across samples, with cancer-associated attributions shown in blue and healthy-associated attributions shown in salmon. Bars are mirrored around zero to facilitate visual comparison between classes. The classifier assigns high cancer-associated attribution to genes including FOLR1, ELF3, KRT19, and MUC1, suggesting that GiG identifies discriminative molecular signals consistent with epithelial and tumor-associated transcriptional programs in circulating cfRNA.}
\label{fig:Rareseq_binary_IG}
\end{figure}

\begin{figure}[pb!]
\centering
\includegraphics[width=\linewidth,height=\textheight,keepaspectratio]{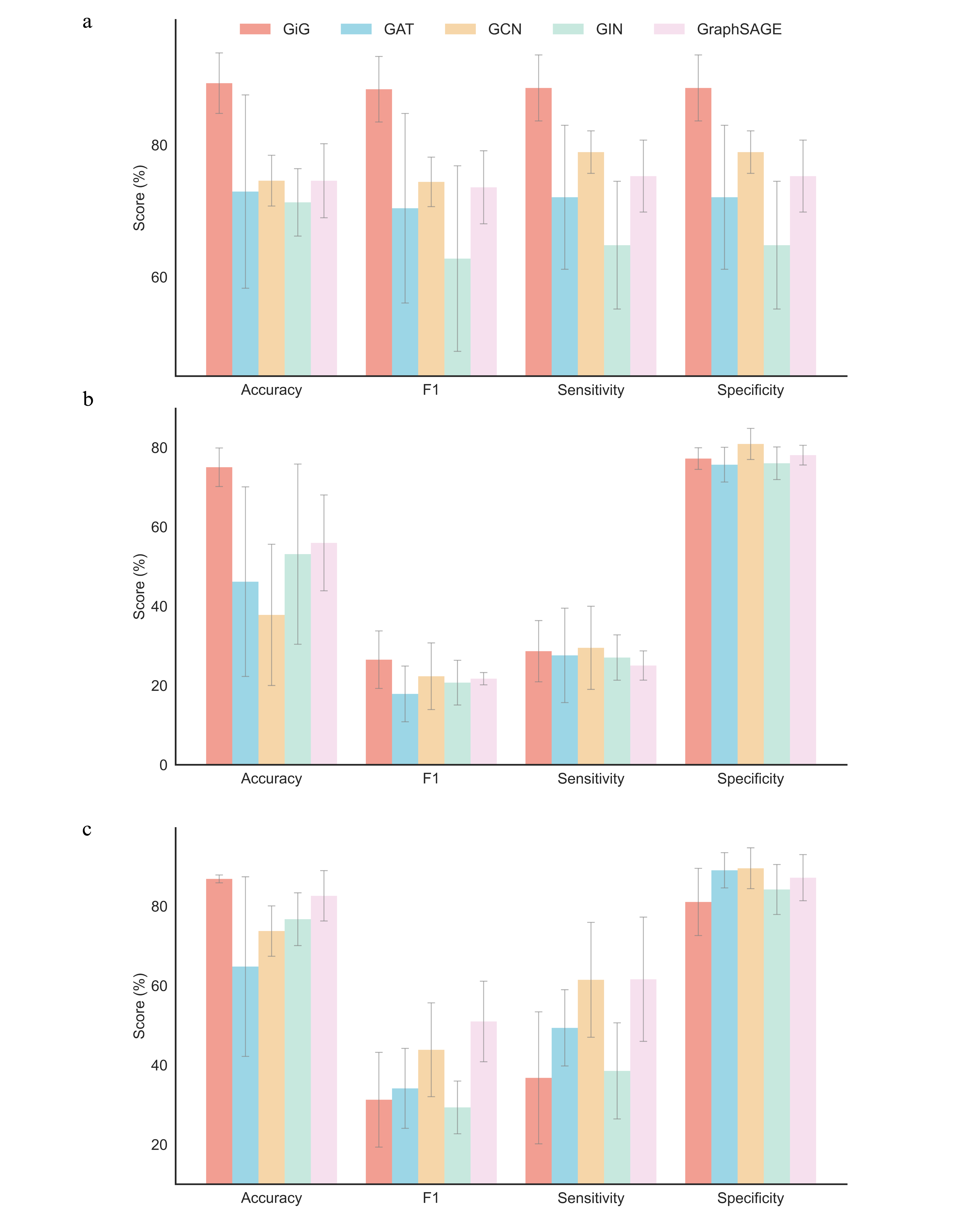}
\caption{\textbf{RARE-Seq binary, stage, and subtype baseline performance.}
Cross-validation performance of GiG compared with node-level GNN baselines across three RARE-Seq classification tasks. a, Binary cancer-versus-healthy classification. b, Cancer stage classification. c, Cancer subtype classification. Bars indicate mean accuracy, macro-F1, sensitivity, and specificity across folds, and error bars indicate variability across folds. GiG shows the strongest overall performance in the binary detection task and maintains competitive performance across stage and subtype prediction, highlighting the benefit of incorporating patient-specific pathway graph structure into cfRNA-based classification.}
\label{fig:Rareseq_all_task_baselines}
\end{figure}

\begin{figure}[pb!]
\centering
\includegraphics[width=\linewidth]{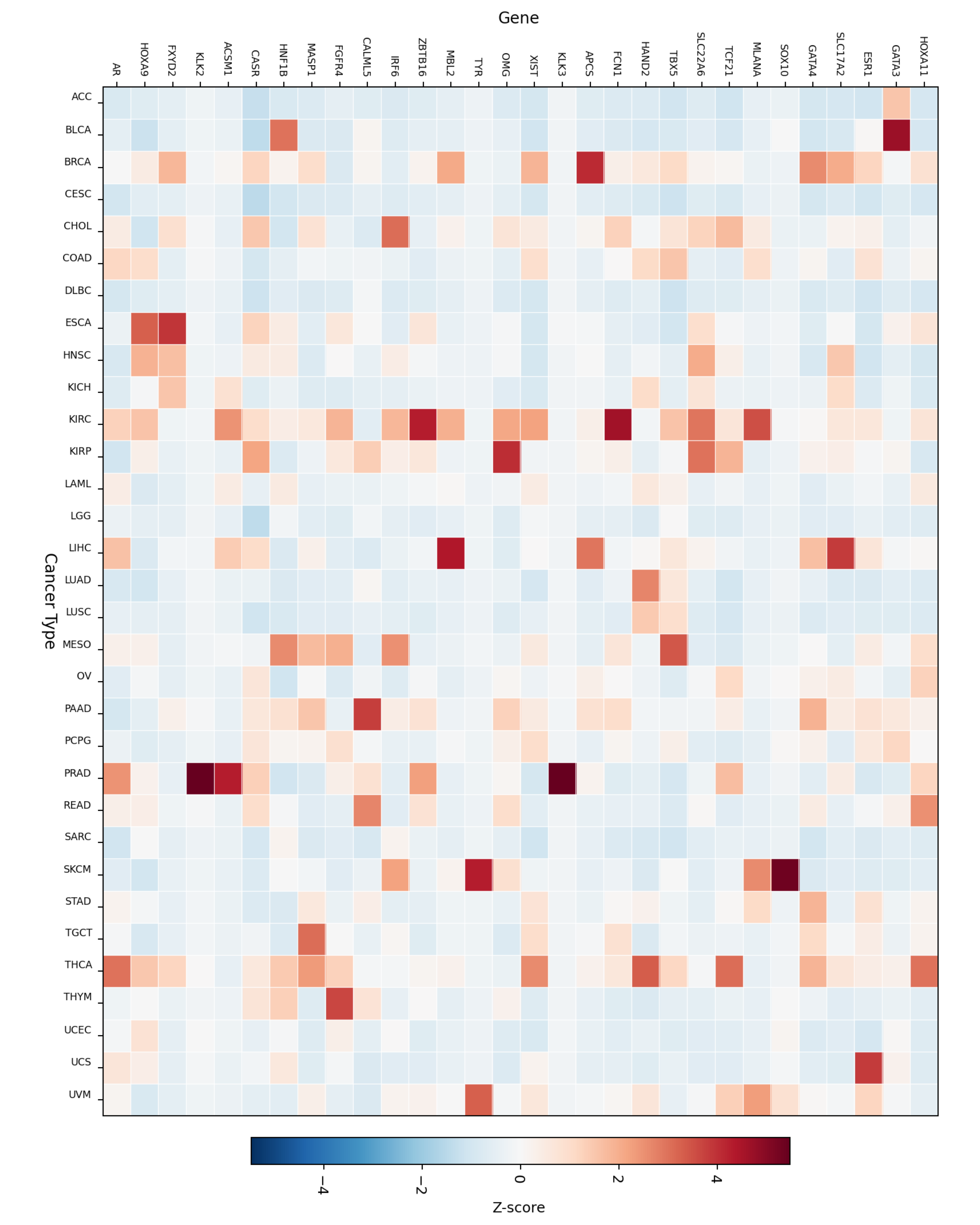}
\caption{\textbf{Pan-cancer Integrated Gradients attribution heatmap.}
Z-scored class-wise Integrated Gradients (IG) attribution scores for the top-ranked genes in the pan-cancer classification task. Rows represent genes and columns represent cancer types. Red indicates higher relative attribution for a given cancer class, while blue indicates lower relative attribution. The heatmap shows that GiG assigns class-dependent importance to distinct gene subsets rather than relying on a single shared transcriptional signature across all tumor types.}
\label{fig:supp_Pancancer_full_heatmap}
\end{figure}

\begin{figure}[pb!]
\centering
\includegraphics[width=\linewidth]{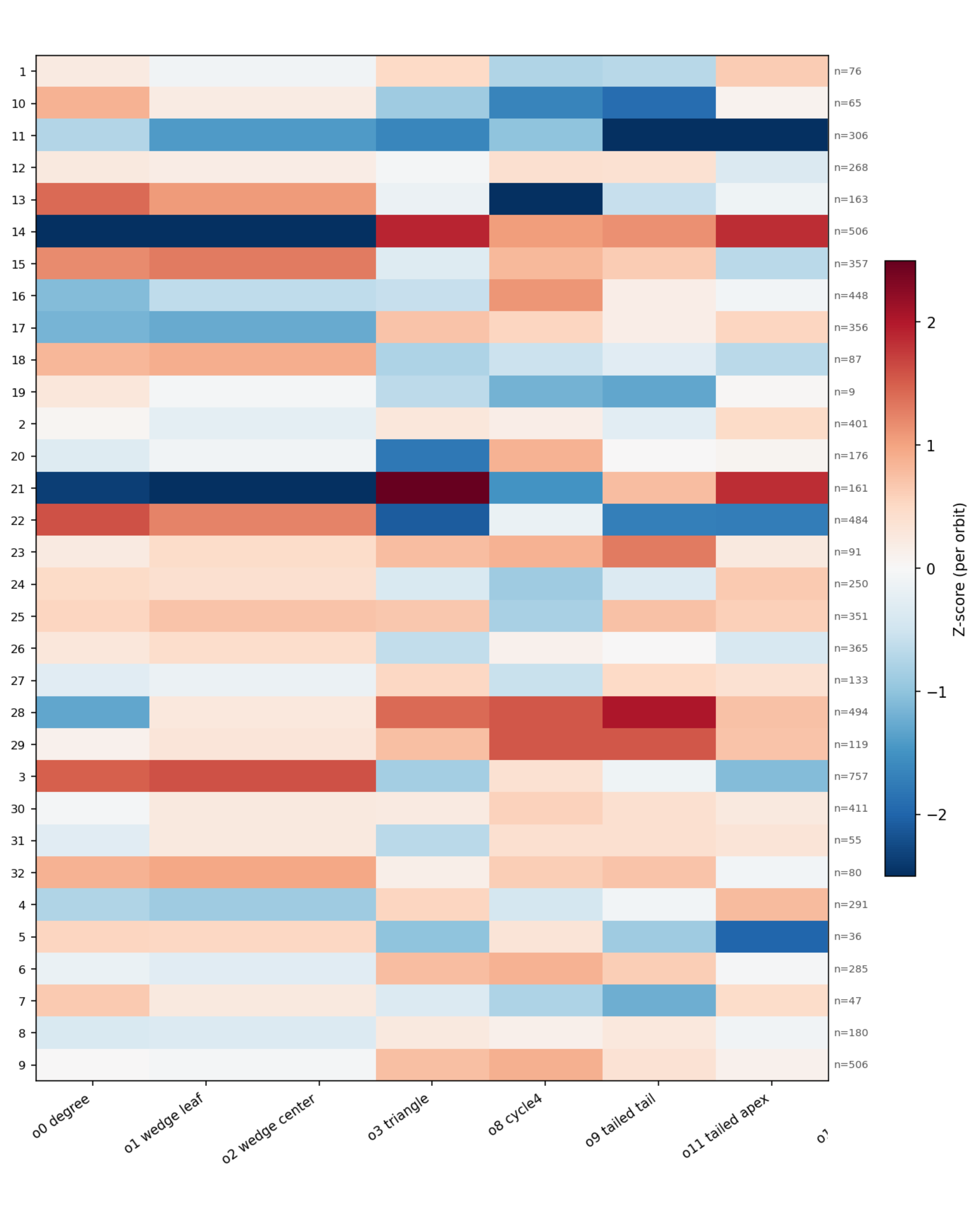}
\caption{\textbf{Cancer type–specific graphlet orbit signatures in pan-cancer pathway graphs.}
Mean ORCA-derived graphlet orbit signatures for real patient-specific pan-cancer pathway graphs, aggregated by cancer type and z-scored within each orbit. Columns correspond to selected 4-node graphlet orbits, including degree, wedge, triangle, cycle, tailed-triangle, and clique-related positions. Each row represents a cancer type, with sample size shown on the right. The heatmap demonstrates that patient-specific pathway graphs differ not only in gene composition but also in local network topology across cancer classes.}
\label{fig:pancancer_cancer_type_heatmap}
\end{figure}

\begin{figure}
\centering
\includegraphics[width=\linewidth]{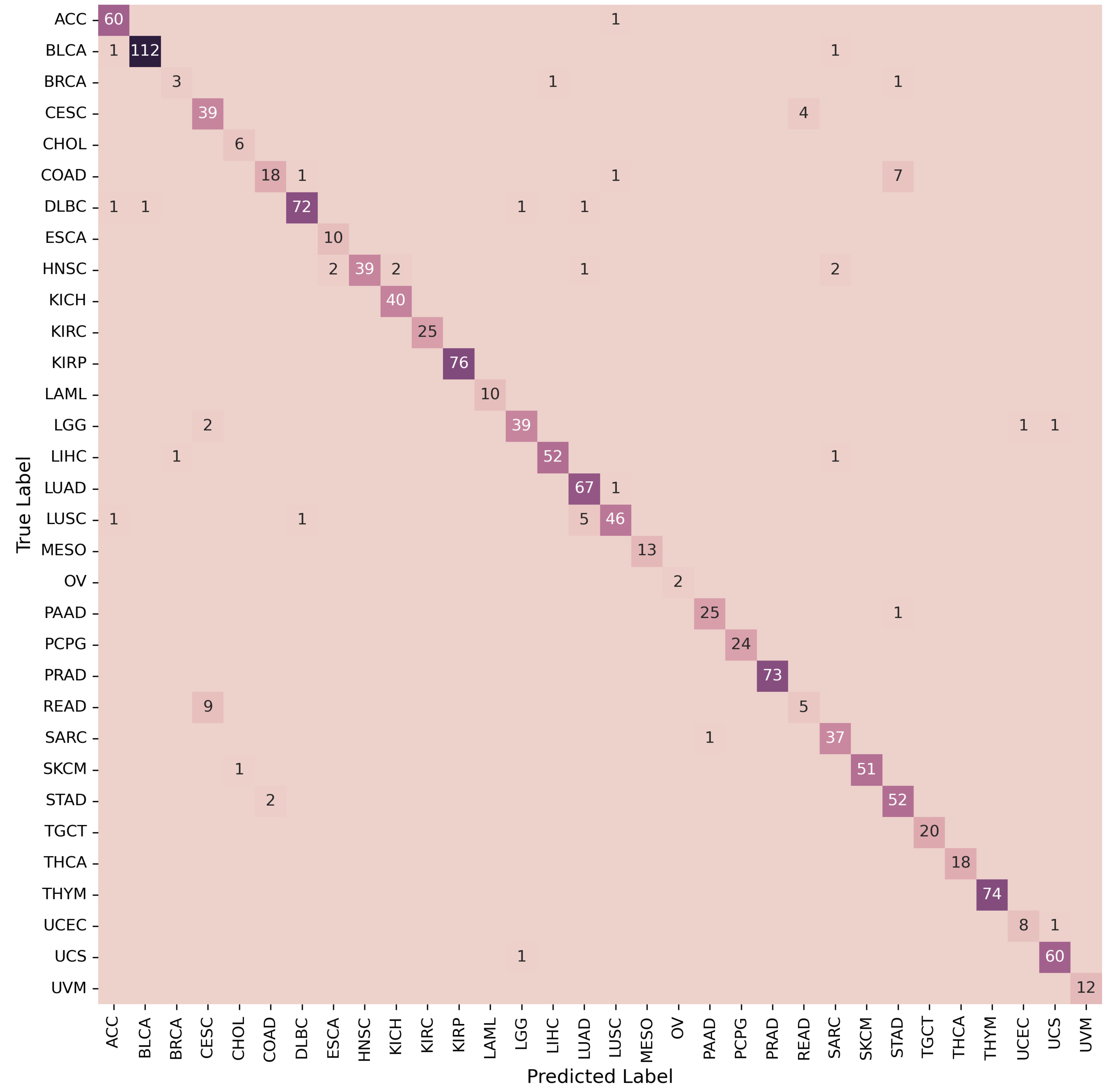}
\caption{\textbf{Pan-cancer full-gene confusion matrix.}
Confusion matrix for the pan-cancer classification task using the full-gene input representation. Rows indicate true cancer labels and columns indicate predicted labels, with values corresponding to sample counts. Darker diagonal entries represent correctly classified samples, while off-diagonal entries indicate misclassifications between cancer types. The strong diagonal structure shows that the model accurately distinguishes most cancer classes, with limited confusion concentrated among a small subset of tumor types.}
\label{fig:Pancancer full confusion matrix}
\end{figure}

\begin{figure}
\centering
\includegraphics[width=\linewidth]{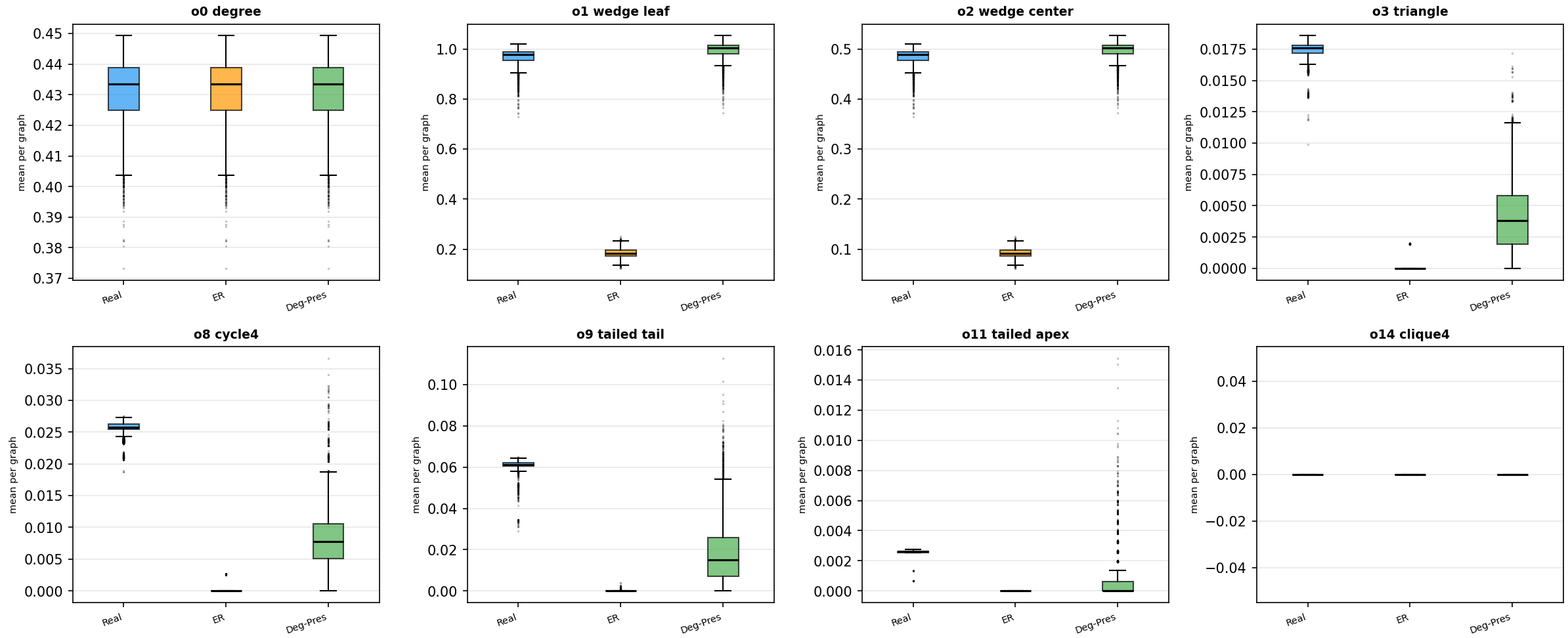}
\caption{\textbf{Pan-cancer pathway graph topology compared with random graph controls:}
Distributions of selected graphlet orbit features in real pan-cancer pathway graphs compared with Erdős–Rényi and degree-preserving random controls. Each boxplot shows the mean orbit count per graph for a specific local topology. Real pathway graphs show distinct enrichment patterns across wedge, triangle, cycle, and tailed-triangle orbits, indicating that the observed patient-specific graphs retain non-random pathway structure.}
\label{fig:pancancer_orbit_distributions}
\end{figure}

\begin{figure}[pb!]
\centering
\includegraphics[width=\linewidth]{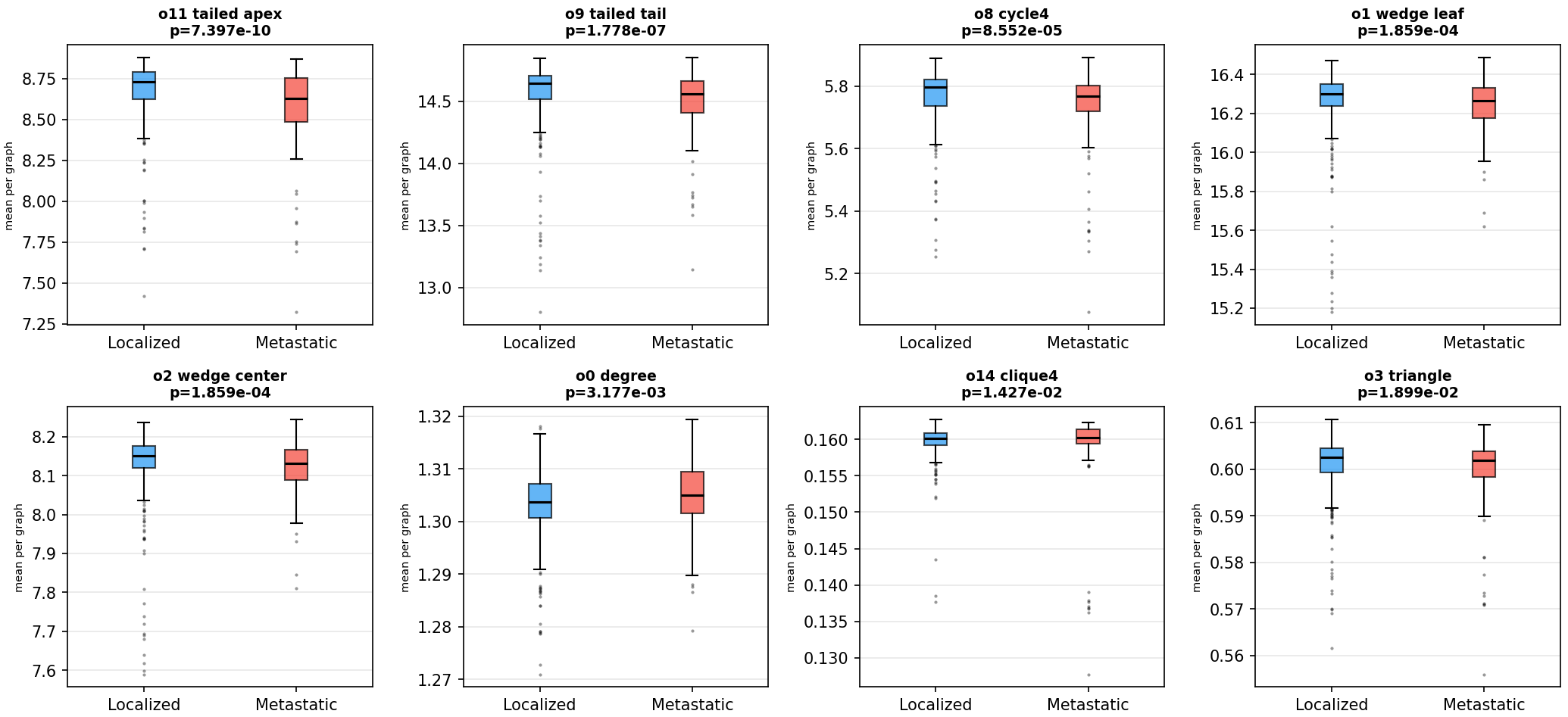}
\caption{\textbf{Graphlet orbit differences between localized and metastatic prostate cancer.}
Top graphlet orbit features distinguishing localized and metastatic prostate cancer pathway graphs, ranked by statistical significance. Boxplots compare mean orbit counts per graph between clinical groups for selected ORCA-derived orbit positions. Differences across tailed-triangle, wedge, cycle, degree, clique, and triangle-related orbits suggest that disease progression is associated with measurable changes in local pathway graph organization.}
\label{fig:prostate_cancer_label_comparison}
\end{figure}

\begin{figure}[pb!]
\centering
\includegraphics[width=\linewidth]{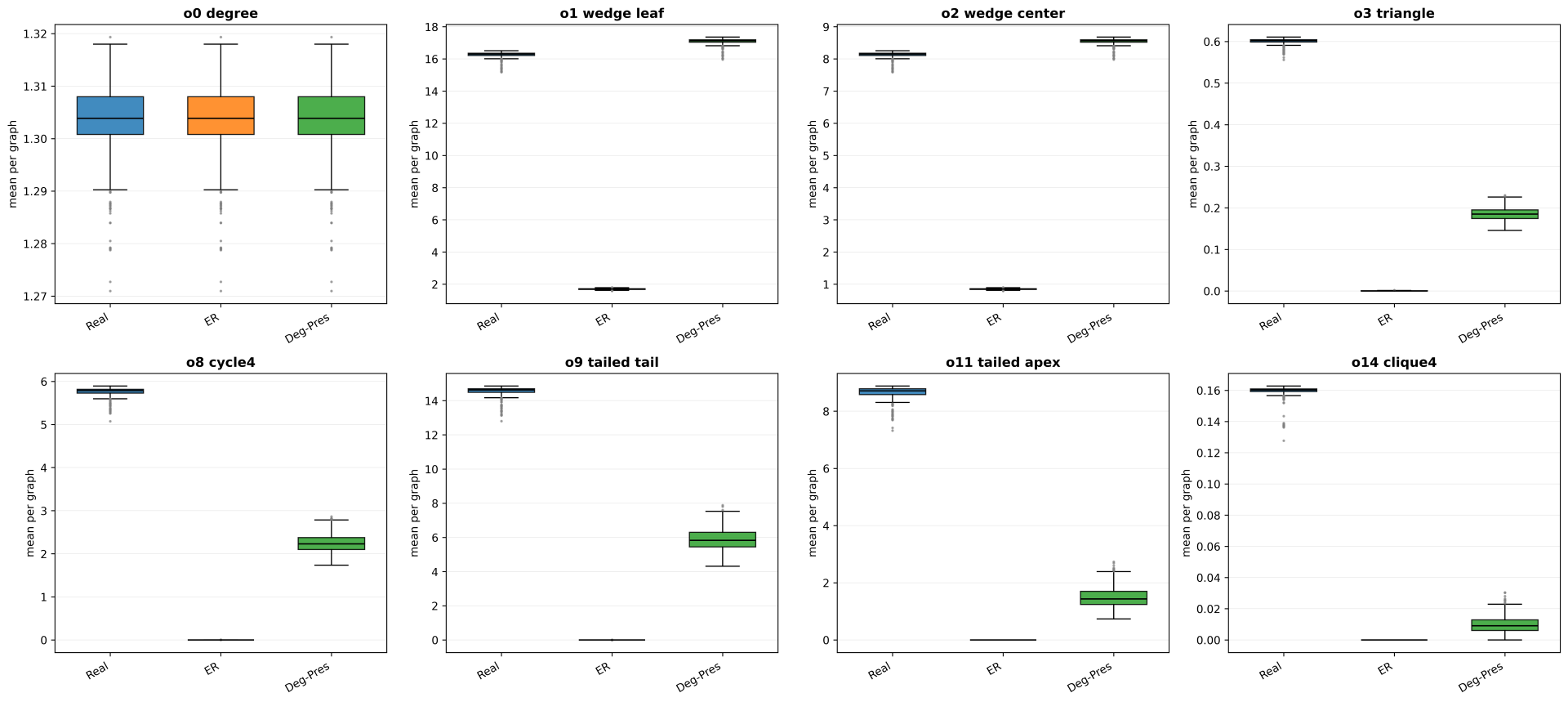}
\caption{\textbf{Prostate cancer pathway graph topology compared with random graph controls.}
Distributions of selected graphlet orbit features in real prostate cancer patient-specific pathway graphs compared with Erdős–Rényi, degree-preserving, and fully connected graph controls. Real pathway graphs display orbit distributions that differ strongly from random and fully connected controls, supporting that curated pathway topology contributes structured, biologically constrained graph organization rather than arbitrary connectivity.}
\label{fig:prostate_cancer_orbit_distributions}
\end{figure}

\begin{figure}[pb!]
\centering
\includegraphics[width=\linewidth]{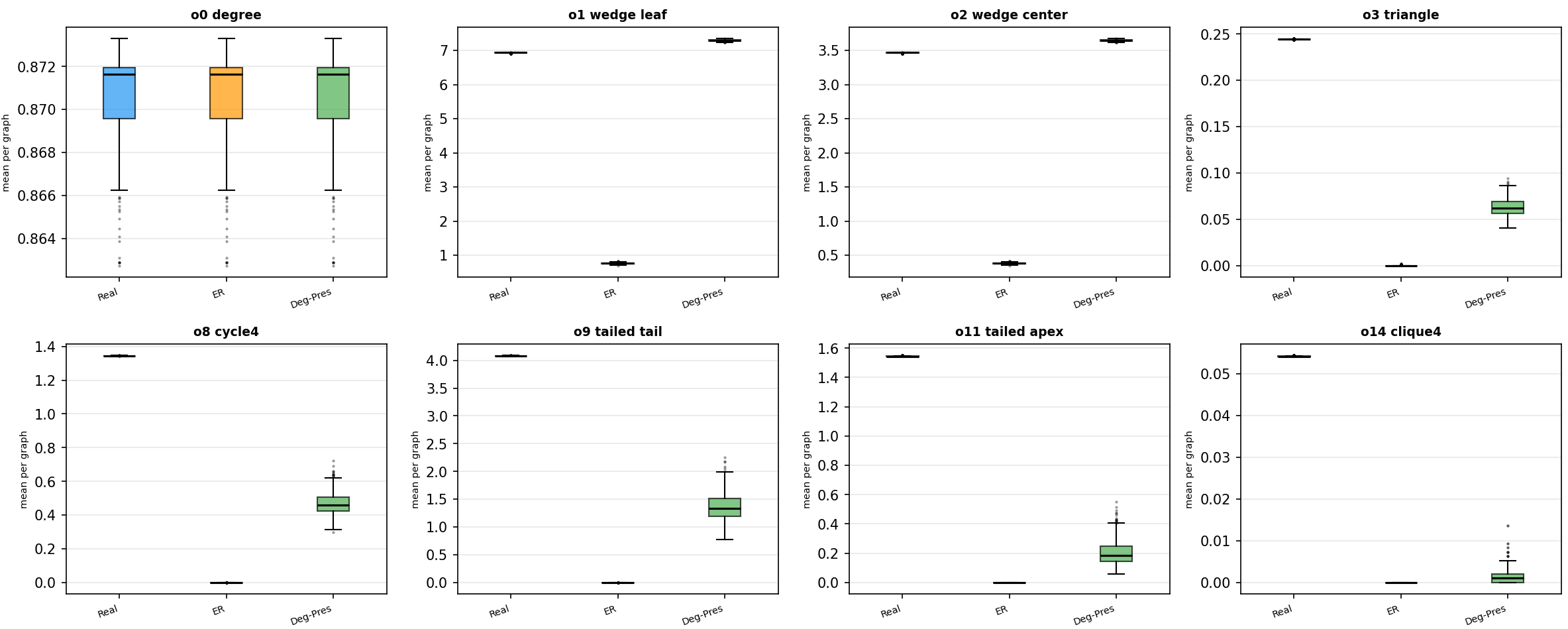}
\caption{\textbf{RareSeq pathway graph topology compared with random graph controls.}
Distributions of selected graphlet orbit features in real RareSeq patient-specific pathway graphs compared with Erdős–Rényi and degree-preserving random controls. Each boxplot shows the mean orbit count per graph for a specific local topology. Real pathway graphs retain elevated counts of closed-motif orbits including triangles (o3), 4-cycles (o8), tailed triangles (o9, o11), and 4-cliques (o14). Erdős–Rényi rewiring collapses these orbits to near zero. Degree-preserving rewiring recovers part of the closed-motif structure but remains well below real-graph values. The degree sequence alone does not reproduce the local clustering present in curated pathway graphs.}
\label{fig:rareseq_orbit_distributions}
\end{figure}

\begin{figure}[pb!]
\centering
\includegraphics[width=\linewidth]{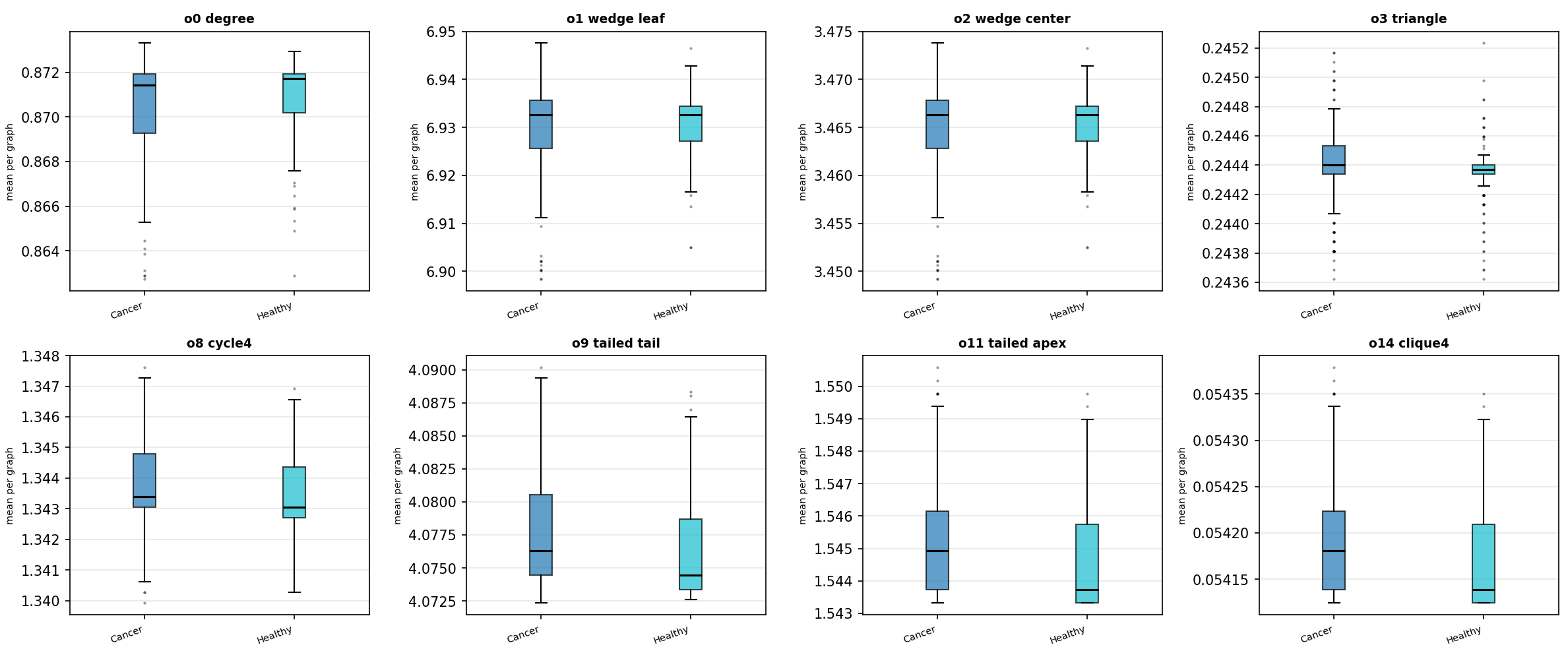}
\caption{\textbf{Graphlet orbit differences between cancer and healthy RARE-Seq binary pathway graphs.} Boxplots compare selected ORCA-derived graphlet orbit features between cancer and healthy patient-specific pathway graphs in the RARE-Seq binary classification task. Each panel shows the mean orbit count per graph for a distinct local topology, including degree, wedge, triangle, cycle, tailed-triangle, and clique-related orbit positions. The distributions indicate that cancer and healthy samples exhibit measurable differences in local pathway graph organization.}
\label{fig:rareseq_binary_orbit_label_comparison}
\end{figure}

\begin{figure}[pb!]
\centering
\includegraphics[width=\linewidth]{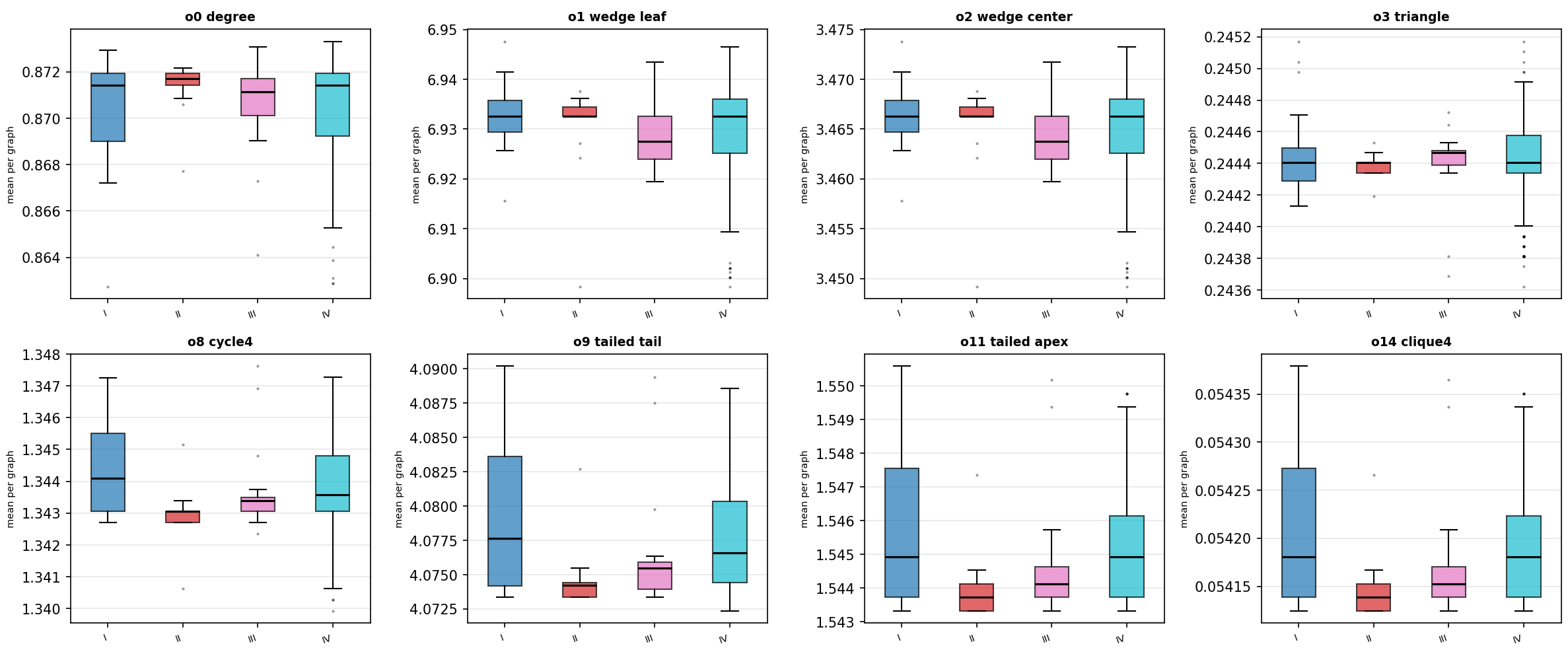}
\caption{\textbf{Graphlet orbit differences across RARE-Seq cancer stages.} Boxplots compare selected ORCA-derived graphlet orbit features across cancer stages in RARE-Seq patient-specific pathway graphs. Each panel shows the mean orbit count per graph for a distinct local topology, including degree, wedge, triangle, cycle, tailed-triangle, and clique-related orbit positions. The distributions suggest that local pathway graph organization varies across stage groups, reflecting stage-associated differences in patient-specific pathway structure.}
\label{fig:rareseq_stage_orbit_label_comparison}
\end{figure}

\begin{figure}[pb!]
\centering
\includegraphics[width=\linewidth]{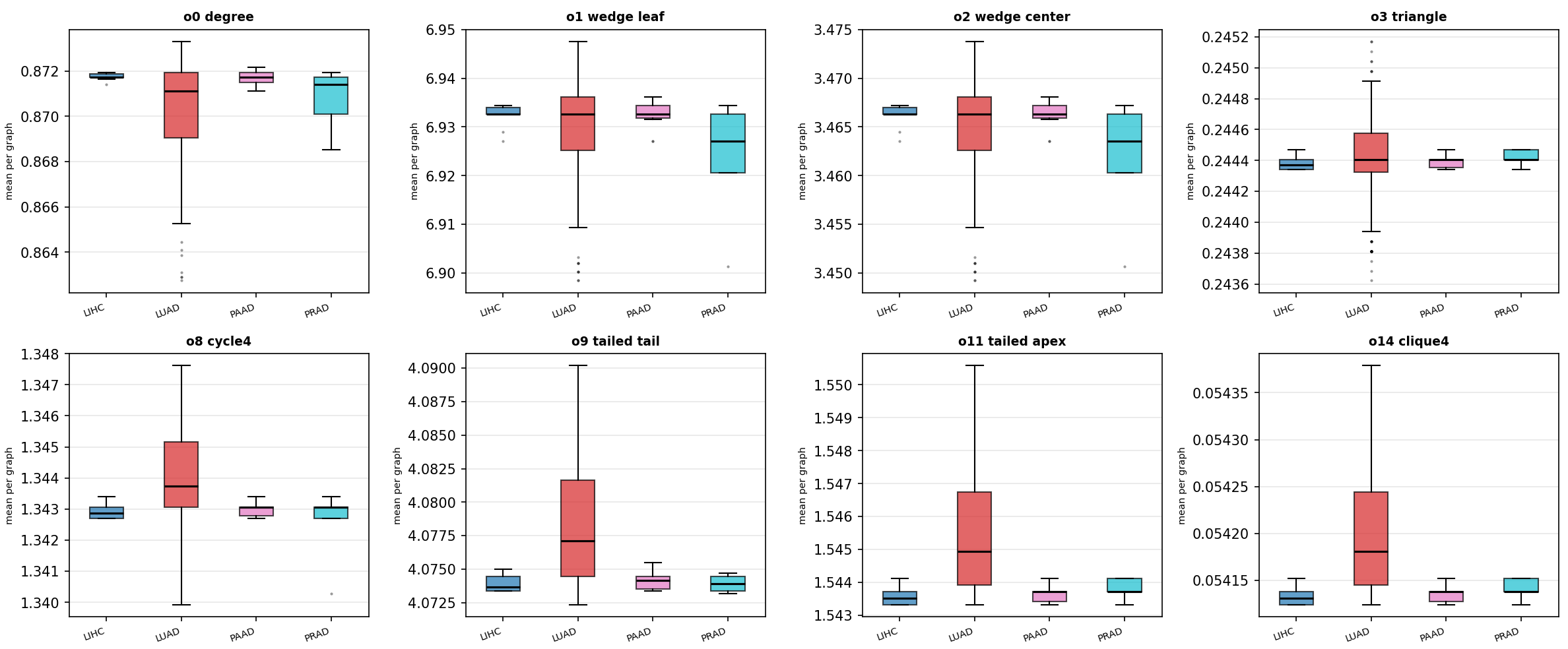}
\caption{\textbf{Graphlet orbit differences across RARE-Seq cancer subtypes.} Boxplots compare selected ORCA-derived graphlet orbit features across cancer subtypes in RARE-Seq patient-specific pathway graphs. Each panel shows the mean orbit count per graph for a distinct local topology, including degree, wedge, triangle, cycle, tailed-triangle, and clique-related orbit positions. The distributions indicate subtype-dependent variation in local pathway graph structure, consistent with differences in molecular organization across cancer types.}
\label{fig:rareseq_subtype_orbit_label_comparison}
\end{figure}

\end{document}